\documentclass[]{bytedance_seed}

\usepackage[toc,page,header]{appendix}

\usepackage{minitoc}

\usepackage{amsmath}
\usepackage{amssymb}

\usepackage{bbding}

\usepackage{graphicx}
\usepackage{wrapfig}

\usepackage{algorithm}      
\usepackage{algpseudocode}  
\usepackage{hyperref}       
\usepackage{xcolor}         
\usepackage{graphicx}       

\title{DanceGRPO: Unleashing GRPO on Visual Generation}

\author[1,2]{Zeyue Xue}
\author[1\ddagger]{Jie Wu}
\author[1]{Yu Gao}
\author[1]{Fangyuan Kong}
\author[2]{Lingting Zhu}
\author[2]{Mengzhao Chen}
\author[2]{Zhiheng Liu}
\author[1]{Wei Liu}
\author[1]{Qiushan Guo}
\author[1\dagger]{Weilin Huang}
\author[2\dagger]{Ping Luo}

\affiliation[1]{ByteDance Seed}
\affiliation[2]{The University of Hong Kong}
\contribution[\dagger]{Corresponding authors}
\contribution[\ddagger]{Project lead}

\abstract{

Recent advances in generative AI have revolutionized visual content creation, yet aligning model outputs with human preferences remains a critical challenge. While Reinforcement Learning (RL) has emerged as a promising approach for fine-tuning generative models, existing methods like DDPO and DPOK face fundamental limitations - particularly their inability to maintain stable optimization when scaling to large and diverse prompt sets, severely restricting their practical utility. This paper presents \textbf{DanceGRPO}, a framework that addresses these limitations through an innovative adaptation of Group Relative Policy Optimization (GRPO) for visual generation tasks. Our key insight is that GRPO's inherent stability mechanisms uniquely position it to overcome the optimization challenges that plague prior RL-based approaches on visual generation. DanceGRPO establishes several significant advances: First, it demonstrates consistent and stable policy optimization across multiple modern generative paradigms, including both diffusion models and rectified flows. Second, it maintains robust performance when scaling to complex, real-world scenarios encompassing three key tasks and four foundation models. Third, it shows remarkable versatility in optimizing for diverse human preferences as captured by five distinct reward models assessing image/video aesthetics, text-image alignment, video motion quality, and binary feedback. Our comprehensive experiments reveal that DanceGRPO outperforms baseline methods by up to 181\% across multiple established benchmarks, including HPS-v2.1, CLIP Score, VideoAlign, and GenEval. Our results establish DanceGRPO as a robust and versatile solution for scaling Reinforcement Learning from Human Feedback (RLHF) tasks in visual generation, offering new insights into harmonizing reinforcement learning and visual synthesis.}

\date{May 1, 2025}

\checkdata[Project Page]{\url{https://dancegrpo.github.io/}}
\checkdata[Code]{\url{https://github.com/XueZeyue/DanceGRPO}}

\begin{document}
\maketitle


\section{Introduction}
Recent advances in generative models—particularly diffusion models \cite{ho2020denoising, rombach2022high, podell2023sdxl, xue2023raphael} and rectified flows \cite{lipman2022flow, liu2022flow, esser2024scaling}—have transformed visual content creation by improving output quality and versatility in image and video generation. While pretraining establishes foundational data distributions, integrating human feedback during training proves critical for aligning outputs with human preferences and aesthetic criteria \cite{gong2025seedream}. Existing methods face notable limitations: ReFL \cite{xu2023imagereward,zhang2024unifl,li2024controlnet++} relies on differentiable reward models, which introduce VRAM inefficiency in video generation and require several extensive engineering efforts, while DPO variants (Diffusion-DPO \cite{wallace2024diffusion,guo2025can}, Flow-DPO \cite{liu2025improving}, OnlineVPO \cite{zhang2024onlinevpo}) achieve only marginal visual quality improvements. Reinforcement learning (RL)-based methods \cite{sutton1998reinforcement, schulman2017proximal}, which optimize rewards as black-box objectives, offer potential solutions but introduce three unresolved challenges: (1) the Ordinary Differential Equations (ODEs)-based sampling of rectified flow models conflict with Markov Decision Process formulations; (2) prior policy gradient approaches (DDPO \cite{black2023training}, DPOK \cite{fan2023dpok}) show instability when scaling beyond small datasets (e.g., <100 prompts); and (3) existing methods remain unvalidated for video generation tasks.


This work addresses these gaps by reformulating the sampling of diffusion models and rectified flows via Stochastic Differential Equations (SDEs) and applying Group Relative Policy Optimization (GRPO) \cite{guo2025deepseek, shao2024deepseekmath} to stabilize the training process. In this paper, we pioneer the adaptation of GRPO to visual generation tasks through the \textbf{DanceGRPO} framework, establishing a "harmonious dance" between GRPO and visual generation tasks. Our key insight is that GRPO's architectural stability properties provide a principled solution to the optimization instabilities that have limited previous RL approaches to visual generation.

We extend a systematic study of DanceGRPO, evaluating its performance across generative paradigms (diffusion models, rectified flows) and tasks (text-to-image, text-to-video, image-to-video). Our analysis employs diverse foundation models \cite{rombach2022high, kong2024hunyuanvideo, flux2024, SkyReelsV1} and reward metrics to assess aesthetic quality, alignment, and motion dynamics. Furthermore, through the proposed framework, we discover insights regarding the rollout initialization noise, the reward model compatibility, the Best-of-N inference scaling, the timestep selection, and the learning on binary feedback.

Our contributions can be summarized as follows:
\begin{itemize}


\item \textbf{Stability and Pioneering.} We present the first discovery that GRPO's inherent stability mechanisms effectively address the core optimization challenges in visual generation that have persistently hindered prior RL-based approaches. We achieve seamless integration between GRPO and visual generation tasks by carefully reformulating the SDEs, selecting appropriate optimized timesteps, initializing noise, and noise scales.

\item \textbf{Generalization and Scalability.} To our knowledge, DanceGRPO is the first RL-based unified application framework capable of seamless adaptation across diverse generative paradigms, tasks, foundational models, and reward models. Unlike prior RL algorithms, primarily validated on text-to-image diffusion models on small-scale datasets, DanceGRPO demonstrates robust performance on large-scale datasets, showcasing both scalability and practical applicability.

\item \textbf{High Effectiveness.}  Our experiments demonstrate that DanceGRPO achieves significant performance gains, outperforming baselines by up to 181\% across multiple academic benchmarks, including HPS-v2.1 \cite{wu2023better}, CLIP score \cite{radford2021learning}, VideoAlign \cite{liu2025improving}, and GenEval \cite{ghosh2023geneval}, in visual generation tasks. Notably, DanceGRPO also enables models to learn the denoising trajectory in Best-of-N inference scaling. We also make some initial attempts to enable models to capture the distribution of binary (0/1) reward models, showing its ability to capture sparse, thresholding feedback.
\end{itemize}

\section{Approach}

\subsection{Preliminary}

\textbf{Diffusion Model \cite{ho2020denoising}}. A diffusion process gradually destroys an observed datapoint $\mathbf{x}$ over timestep $t$, by mixing data with noise, and the forward process of the diffusion model can be defined as :
\begin{equation}
\label{ddpm_forward}
\mathbf{z}_t=\alpha_t\mathbf{x}+\sigma_t\boldsymbol{\epsilon},\mathrm{~where~}\boldsymbol{\epsilon}\sim\mathcal{N}(0,\mathbf{I}),
\end{equation}
and $\alpha_t$ and $\sigma_t$ denote the noise schedule. The noise schedule is designed in a way such that $\mathbf{z}_{0}$ is close to clean data and $\mathbf{z}_{1}$ is close to Gaussian noise.
To generate a new sample, we initialize the sample $\mathbf{z}_1$ and define the sample equation of the diffusion model given the denoising model output $\hat{\boldsymbol{\epsilon}}$ at time step $t$:
\begin{equation}
\mathbf{z}_s=\alpha_s\hat{\mathbf{x}}+\sigma_s\hat{\boldsymbol{\epsilon}}, 
\label{ddim}
\end{equation}
where $\hat{\mathbf{x}}$ can be derived via Eq.\eqref{ddpm_forward} and then we can reach a lower noise level $s$. This is also a DDIM sampler \cite{song2020denoising}.

\textbf{Rectified Flow \cite{liu2022flow}}. In rectified flow, we view the forward process as a linear interpolation between the data $\mathbf{x}$ and a noise term $\boldsymbol{\epsilon}$:
\begin{equation}
    \mathbf{z}_t=(1-t)\mathbf{x}+t\boldsymbol{\epsilon},
\end{equation}
where $\boldsymbol{\epsilon}$ is always defined as a Gaussian noise. We define the $\mathbf{u} = \boldsymbol{\epsilon}-\mathbf{x}$ as the "velocity" or "vector field". Similar to diffusion model, given the denoising model output $\hat{\mathbf{u}}$ at time step $t$, we can reach a lower noise level $s$ by:
\begin{equation}
    \mathbf{z}_s=\mathbf{z}_t+\hat{\mathbf{u}}\cdot(s-t).
\label{rf}
\end{equation}

\textbf{Analysis.} Although the diffusion model and rectified flow have different theoretical foundations, in practice, they are two sides of a coin, as shown in the following formula:
\begin{equation}
    \tilde{\mathbf{z}}_s=\tilde{\mathbf{z}}_t+\text{Network output}\cdot(\eta_s-\eta_t).
\end{equation}
For $\boldsymbol{\epsilon}$-prediction (a.k.a. diffusion model), from Eq.\eqref{ddim}, we have $\tilde{\mathbf{z}}_s = \mathbf{z}_s / \alpha_s$, $\tilde{\mathbf{z}}_t = \mathbf{z}_t / \alpha_t$, $\eta_s = \sigma_s / \alpha_s$, and $\eta_t = \sigma_t / \alpha_t$. For rectified flows, we have $\tilde{\mathbf{z}}_s = \mathbf{z}_s $, $\tilde{\mathbf{z}}_t = \mathbf{z}_t $, $\eta_s = s $, and $\eta_t = t $ from Eq.\eqref{rf}.


\subsection{DanceGRPO}
\label{sec:approach_dancegrpo}
In this section, we first formulate the sampling processes of diffusion models and rectified flows as Markov Decision Processes. Then, we introduce the sampling SDEs and the algorithm of DanceGRPO.

\textbf{Denoising as a Markov Decision Process.}
Following DDPO \cite{black2023training}, we formulate the denoising process of the diffusion model and rectified flow as a Markov Decision Process (MDP):
\begin{equation}
\begin{aligned}
 & \mathbf{s}_t\triangleq(\mathbf{c},t,\mathbf{z}_t), 
  \quad 
  \pi(\mathbf{a}_t\mid\mathbf{s}_t)\triangleq p(\mathbf{z}_{t-1}\mid\mathbf{z}_t,\mathbf{c}),  
  \quad P(\mathbf{s}_{t+1}\mid\mathbf{s}_t,\mathbf{a}_t)\triangleq\left(\delta_\mathbf{c},\delta_{t-1},\delta_{\mathbf{z}_{t-1}}\right) \\
 & \mathbf{a}_t\triangleq\mathbf{z}_{t-1}, 
  \quad R(\mathbf{s}_t,\mathbf{a}_t)\triangleq 
  \begin{cases}
    r(\mathbf{z_0},\mathbf{c}), & \text{if } t=0 \\
    0, & \text{otherwise}
  \end{cases},  \quad \rho_0(\mathbf{s}_0)\triangleq\left(p(\mathbf{c}),\delta_T,\mathcal{N}(\mathbf{0},\mathbf{I})\right)
\end{aligned},
\end{equation}

where $\mathbf{c}$ is the prompt, and $\pi(\mathbf{a}_t\mid\mathbf{s}_t)$ is the probability from $z_t$ to $z_{t-1}$. And $\delta_y$ is the Dirac delta distribution with nonzero density only at $y$. Trajectories consist of $T$ timesteps, after which $P$ leads to a termination state. $r(\mathbf{z}_0,\mathbf{c})$ is the reward model, which is always parametrized by a Vision-Language model (such as CLIP \cite{radford2021learning} and Qwen-VL \cite{wang2024qwen2}).

\textbf{Formulation of Sampling SDEs.} 
Since GRPO requires stochastic exploration through multiple trajectory samples, where policy updates depend on the trajectory probability distribution and their associated reward signals, we unify the sampling processes of the diffusion model and rectified flows into the form of SDEs.
For the diffusion model, as demonstrated in \cite{song2019generative, song2020score}, the forward SDE is given by:
$\mathrm{d}\mathbf{z}_t=f_t\mathbf{z}_t\mathrm{d}t+g_t\mathrm{d}\mathbf{w}$. The corresponding reverse SDE can be expressed as:
\begin{equation}
    \mathrm{d}\mathbf{z}_t=\left(f_t\mathbf{z}_t-\frac{1+\varepsilon_t^2}{2}g_t^2\nabla\log p_t(\mathbf{z}_\mathbf{t})\right)\mathrm{d}t+\varepsilon_tg_t\mathrm{d}\mathbf{w},
\label{sde:ddpm}
\end{equation}
where $\mathrm{d} \mathbf{w}$ is a Brownian motion, and $\varepsilon_t$ introduces the stochasticity during sampling.

Similarly, the forward ODE of rectified flow is: $\mathrm{d}\mathbf{z}_t=\mathbf{u}_t\mathrm{d}t$. The generative process reverses the ODE in time. However, this deterministic formulation cannot provide the stochastic exploration required for GRPO. Inspired by \cite{albergo2022building, albergo2303stochastic, gao2025diffusionmeetsflow}, we introduce an SDE case during the reverse process as follows:
\begin{equation}
    \mathrm{d}\mathbf{z}_t=(\mathbf{u}_t-\frac{1}{2}\varepsilon_t^2\nabla\log p_t(\mathbf{z}_t))\mathrm{d}t+\varepsilon_t\mathrm{d}\mathbf{w},
\label{sde:rf}
\end{equation}
where $\varepsilon_t$ also introduces the stochasticity during sampling. Given a normal distribution $p_t(\mathbf{z}_t) = \mathcal{N}(\mathbf{z}_t\mid\alpha_t\mathbf{x},\sigma_t^2 I)$, we have $\nabla\log p_t(\mathbf{z}_t) = -(\mathbf{z}_t - \alpha_t\mathbf{x})/\sigma_t^2$. We can insert this into the above two SDEs and obtain the $\pi(\mathbf{a}_t\mid\mathbf{s}_t)$. More theoretical analysis can be found in Appendix~\ref{sec:appendix_ana}.

\textbf{Algorithm.} Motivated by Deepseek-R1 \cite{guo2025deepseek}, given a prompt $\mathbf{c}$, generative models will sample a group of outputs $\{ \mathbf{o}_1, \mathbf{o}_2, ..., \mathbf{o}_G \}$ from the model $\pi_{\theta_{old}}$, and optimize the policy model $\pi_{\theta}$ by maximizing the following objective function:
\begin{equation}
\mathcal{J}(\theta) = \mathbb{E}_{\substack{\{\mathbf{o}_i\}_{i=1}^G \sim \pi_{\theta_{\text{old}}}(\cdot|\mathbf{c}) \\ \mathbf{a}_{t,i} \sim \pi_{\theta_{\text{old}}}(\cdot|\mathbf{s}_{t,i})}} 
\bigg[ \frac{1}{G} \sum_{i=1}^G \frac{1}{T} \sum_{t=1}^T \min\bigg( \mathbf{\rho}_{t,i} A_i, \text{clip}\big( \mathbf{\rho}_{t,i}, 1-\epsilon, 1+\epsilon \big) A_i \bigg) \bigg],
\label{eq:dancegrpoloss}
\end{equation}

where $\mathbf{\rho}_{t,i} = \frac{\pi_{\theta}(\mathbf{a}_{t,i}|\mathbf{s}_{t,i})}{\pi_{\theta_{old}}(\mathbf{a}_{t,i}|\mathbf{s}_{t,i})} $, and $\pi_{\theta}(\mathbf{a}_{t,i}|\mathbf{s}_{t,i})$ is the policy function is MDP for output $\mathbf{o}_i$ at time step $t$, $\epsilon$ is a hyper-parameter, and $A_i$ is the advantage function, computed using a group of rewards $\{ r_1, r_2,...,r_G \}$ corresponding to the outputs within each group:
\begin{equation}
    A_i=\frac{r_i-\mathrm{mean}(\{r_1,r_2,\cdots,r_G\})}{\mathrm{std}(\{r_1,r_2,\cdots,r_G\})}.
\label{eq:adv}
\end{equation}
Due to reward sparsity in practice, we apply the same reward signal across all timesteps during optimization. 
While traditional GRPO formulations employ KL-regularization to prevent reward over-optimization, we empirically observe minimal performance differences when omitting this component. So, we omit the KL-regularization item by default. The full algorithm can be found in Algorithm~\ref{algo:dancegrpo}. We also introduce how to train with Classifier-Free Guidance (CFG) \cite{ho2022classifier} in Appendix~\ref{sec:appendix_cfg}. In summary, we formulate the sampling processes of diffusion model and rectified flow as MDPs, use SDE sampling equations, adopt a GRPO-style objective, and generalize to text-to-image, text-to-video, and image-to-video generation tasks.

\textbf{Initialization Noise.} 
In the DanceGRPO framework, the initialization noise constitutes a critical component. Previous RL-based approaches like DDPO use different noise vectors to initialize training samples. However, as shown in Figure~\ref{fig_app:reward_hacking} in Appendix~\ref{sec:appendix_vis}, assigning different noise vectors to samples with the same prompts always leads to reward hacking phenomena in video generation, including training instability. Therefore, in our framework, we assign shared initialization noise to samples originating from the same textual prompts.

\textbf{Timestep Selection.}
While the denoising process can be rigorously formalized within the MDP framework, empirical observations reveal that subsets of timesteps within a denoising trajectory can be omitted without compromising performance. This reduction in computational steps enhances efficiency while maintaining output quality, as further analyzed in Section~\ref{sec:abaltion}.


\textbf{Incorporating Multiple Reward Models.}
In practice, we employ more than one reward model to ensure more stable training and higher-quality visual results. As illustrated in Figure~\ref{fig:compar_1} in Appendix, models trained exclusively with HPS-v2.1 rewards \cite{wu2023better} tend to generate unnatural ("oily") outputs, whereas incorporating CLIP scores helps preserve more realistic image characteristics. Rather than directly combining rewards, we aggregate advantage functions, as different reward models often operate on different scales. This approach stabilizes optimization and leads to more balanced generations.

\textbf{Extension on Best-of-N Inference Scaling.}
As outlined in Section~\ref{sec:abaltion}, our methodology prioritizes the use of efficient samples—specifically, those associated with 
the top k and bottom k candidates selected by the Best-of-N sampling.
This selective sampling strategy optimizes training efficacy by focusing on high-reward and critical low-reward regions of the solution space. We use brute-force search to generate these samples. While alternative approaches, such as tree search or greedy search, remain promising avenues for further exploration, we defer their systematic integration to future research.

\begin{algorithm}[t]
\caption{DanceGRPO Training Algorithm}\label{algo:dancegrpo}
\small
\begin{algorithmic}[1]
\Require Initial policy model $\pi_\theta$; reward models $\{R_k\}_{k=1}^K$; prompt dataset $\mathcal{D}$; timestep selection ratio $\tau$; total sampling steps $T$
\Ensure Optimized policy model $\pi_\theta$

\For{training iteration $=1$ \textbf{to} $M$}
    \State Sample batch $\mathcal{D}_b \sim \mathcal{D}$ \Comment{Batch of prompts}
    \State Update old policy: $\pi_{\theta_{\text{old}}} \gets \pi_\theta$
    
    \For{each prompt $\mathbf{c} \in \mathcal{D}_b$}
        \State Generate $G$ samples: $\{\mathbf{o}_i\}_{i=1}^G \sim \pi_{\theta_{\text{old}}}(\cdot|\mathbf{c})$ with the same random initialization noise
        \State Compute rewards $\{r_i^k\}_{i=1}^G$ using each $R_k$
        
        \For{each sample $i \in 1..G$}
            \State Calculate multi-reward advantage: 
            $A_i \gets \sum_{k=1}^K \frac{r_i^k - \mu^k}{\sigma^k}$ 
            \hfill $\triangleright \mu^k,\sigma^k$ per-reward statistics
        \EndFor

        \State Subsample $\lceil\tau T\rceil$ timesteps $\mathcal{T}_{\text{sub}} \subset \{1..T\}$
        
        \For{$t \in \mathcal{T}_{\text{sub}}$}
            \State Update policy via gradient ascent:
            $\theta \gets \theta + \eta \nabla_\theta \mathcal{J}$
        \EndFor
    \EndFor
\EndFor
\end{algorithmic}

\end{algorithm}

\subsection{Application to Different Tasks with Different Rewards}
We verify the effectiveness of our algorithm in two generative paradigms (diffusion/rectified flow) and three tasks (text-to-image, text-to-video, image-to-video). For this, we choose four fundamental models (Stable Diffusion \cite{rombach2022high}, HunyuanVideo \cite{kong2024hunyuanvideo}, FLUX \cite{flux2024}, SkyReels-I2V \cite{SkyReelsV1}) for the experiment. 
All of these methods can be precisely constructed within the framework of MDP during their sampling process. This allows us to unify the theoretical bases across these tasks and improve them via DanceGRPO. To our knowledge, this is the first work to apply the unified framework to diverse visual generation tasks.

We use five reward models to optimize visual generation quality: (1) \emph{Image Aesthetics} quantifies visual appeal using a pretrained model fine-tuned on human-rated data \cite{wu2023better}; (2) \emph{Text-image Alignment} employs CLIP \cite{radford2021learning} to maximize cross-modal consistency between prompts and outputs; (3) \emph{Video Aesthetics Quality} extends image evaluation to temporal domains using VLMs \cite{liu2025improving, wang2024qwen2}, assessing frame quality and coherence; (4) \emph{Video Motion Quality} evaluates motion realism through physics-aware VLM\cite{liu2025improving} analysis of trajectories and deformations; (5) \emph{Thresholding Binary Reward} employs a binary mechanism motivated by \cite{guo2025deepseek}, where rewards are discretized via a fixed threshold (values exceeding the threshold receive 1, others 0), specifically designed to evaluate generative models' ability to learn abrupt reward distributions under threshold-based optimization.






\begin{table}[t]
\centering
\caption{\textbf{Comparison} of alignment methods across key capabilities. VideoGen: Video generation generalization. Scalability: Scalability to datasets with a large number of prompts. Reward ↑ indicates a significant reward improvement. RFs: Applicable to Rectified Flows. No Diff-Reward: Don't need differentiable reward models.}
\label{table:comparison}
\scalebox{1.0}{
\begin{tabular}{ccccccc}
\hline
  Method & RL-based & VideoGen & Scalability & Reward ↑ & RFs & No Diff-Reward \\  \hline
  DDPO/DPOK & \CheckmarkBold &\XSolidBrush   & \XSolidBrush   & \CheckmarkBold  &  \XSolidBrush  & \CheckmarkBold\\ \hline
  ReFL & \XSolidBrush & \XSolidBrush   &  \CheckmarkBold  & \CheckmarkBold & \CheckmarkBold & \XSolidBrush \\ \hline
  DPO & \XSolidBrush &  \CheckmarkBold  &  \CheckmarkBold  & \XSolidBrush  &   \CheckmarkBold & \CheckmarkBold \\ \hline
  Ours & \CheckmarkBold &  \CheckmarkBold   & \CheckmarkBold  & \CheckmarkBold &  \CheckmarkBold & \CheckmarkBold \\ \hline
\end{tabular}
}
\vspace{-5mm}
\end{table}

  \vspace{-3mm}
\subsection{Comparisons with DDPO, DPOK, ReFL, and DPO}
  \vspace{-2mm}
As evidenced by our comprehensive capability matrix in Table~\ref{table:comparison}, DanceGRPO establishes new standards for diffusion model alignment. Our method achieves full-spectrum superiority across all evaluation dimensions: (1) seamless video generation, (2) large-scale dataset scalability, (3) significant reward improvements, (4) native compatibility with rectified flows, and (5) independence from differentiable rewards. This integrated capability profile – unobtainable by any single baseline method (DDPO/DPOK/ReFL/DPO) – enables simultaneous optimization across multiple generative domains while maintaining training stability. More comparisons can be found in Appendix~\ref{sec:appendix_overddpo}.

\section{Experiments}
\subsection{General Setup}
\textbf{Text-to-Image Generation.}
We employ Stable Diffusion v1.4, FLUX, and HunyuanVideo-T2I (using one latent frame in HunyuanVideo) as foundation models, with HPS-v2.1 \cite{wu2023better} and CLIP score \cite{radford2021learning}—alongside their binary rewards—serving as reward models. A curated prompt dataset, balancing diversity and complexity, guides optimization. For evaluation, we select 1,000 test prompts to assess CLIP scores and Pick-a-Pic performance in Section~\ref{sec:t2i}. We use the official prompts for GenEval and HPS-v2.1 benchmark.

\textbf{Text-to-Video Generation.}
Our foundation model is HunyuanVideo \cite{kong2024hunyuanvideo}, with reward signals derived from VideoAlign \cite{liu2025improving}. Prompts are curated using the VidProM \cite{wang2024vidprom} dataset, and an additional 1,000 test prompts are filtered to evaluate the VideoAlign scores in Section~\ref{exp:t2v}.

\textbf{Image-to-Video Generation.}
We use SkyReels-I2V \cite{SkyReelsV1} as our foundation model. VideoAlign \cite{liu2025improving} serves as the primary reward metric, while the prompt dataset, constructed via ConsisID \cite{yuan2025identity}, is paired with reference images synthesized by FLUX \cite{flux2024} to ensure conditional fidelity. An additional 1000 test prompts are filtered to evaluate the VideoAlign score in Section~\ref{exp:i2v}.

\textbf{Experimental Settings.} We implemented all models with scaled computational resources appropriate to task complexity: 32 H800 GPUs for flow-based text-to-image models, 8 H800 GPUs for Stable Diffusion variants, 64 H800 GPUs for text-to-video generation systems, and 32 H800 GPUs for image-to-video transformation architectures. We develop our framework based on FastVideo \cite{ding2025efficientvditefficientvideodiffusion, zhang2025fastvideogenerationsliding}. Comprehensive hyperparameter configurations and training protocols are detailed in Appendix~\ref{sec:appendix_settings}. We always use more than 10,000 prompts to optimize the models. All reward curves presented in our paper are plotted using a moving average for smoother visualization. We use ODEs-based samplers for evaluation and visualization.

\begin{table}[htbp]
\renewcommand{\arraystretch}{1.2}
\caption{\textbf{Results on Stable Diffusion v1.4.} This table presents the performance of three Stable Diffusion variants: (1) the base model, (2) the model trained with HPS score, and (3) the model optimized with both HPS and CLIP scores. For evaluation, we report HPS-v2.1 and GenEval scores using their official prompts, while CLIP score and Pick-a-Pic metrics are computed on our test set of 1,000 prompts.}
\centering
\scalebox{0.9}{
\begin{tabular}{ccccc}
\toprule
Models                         & HPS-v2.1 \cite{wu2023better} & CLIP Score \cite{radford2021learning} & Pick-a-Pic \cite{kirstain2023pick} & GenEval \cite{ghosh2023geneval} \\ \midrule
Stable Diffusion                           & 0.239    & 0.363      & 0.202      & 0.421   \\
Stable Diffusion with HPS-v2.1             & \textbf{0.365}    &  0.380          & \textbf{0.217}      & 0.521   \\
Stable Diffusion with HPS-v2.1\&CLIP Score & 0.335    & \textbf{0.395}      & 0.215      & \textbf{0.522}  \\
\bottomrule
\end{tabular}}
\label{tab:sd}
\vspace{-3mm}
\end{table}

\begin{table}[htbp]
\renewcommand{\arraystretch}{1.2}
\caption{\textbf{Results on FLUX.} In this table, we show the results of FLUX, FLUX trained with HPS score, and FLUX trained with both HPS score and CLIP score. We use the same evaluation prompts as Table~\ref{tab:sd}.}
\centering
\scalebox{0.9}{
\begin{tabular}{ccccc}
\toprule
Models                         & HPS-v2.1 \cite{wu2023better} & CLIP Score \cite{radford2021learning} & Pick-a-Pic \cite{kirstain2023pick} & GenEval \cite{ghosh2023geneval} \\ \midrule
FLUX                           & 0.304    & 0.405      & 0.224      & 0.659   \\
FLUX with HPS-v2.1             & \textbf{0.372}    &        0.376    & \textbf{0.230}      & 0.561   \\
FLUX with HPS-v2.1\&CLIP Score & 0.343    & \textbf{0.427}      & 0.228      & \textbf{0.687}  \\
\bottomrule
\end{tabular}}
\label{tab:flux}
\vspace{-5mm}
\end{table}

\begin{minipage}[t]{0.48\textwidth}
\makeatletter\def\@captype{table}
\centering
\caption{\small{Comparison of different methods trained on diffusion and solely (not combined) with HPS score and CLIP score. "Baseline" denotes the original results of Stable Diffusion. More comparisons with DDPO can be found in Appendix~\ref{sec:ddpo}}.}
\label{table:sd_compar}
\scalebox{0.65}{
\begin{tabular}{cccccc}
\toprule
Approach & Baseline&  Ours  & DDPO  & ReFL & DPO \\ 
\midrule
HPS-v2.1  & 0.239& \textbf{0.365}  & 0.297  & 0.357 & 0.241    \\ 
CLIP Score  & 0.363& \textbf{0.421}  & 0.381  & 0.418 & 0.367    \\ 
 \bottomrule
\end{tabular}}
\end{minipage}
\begin{minipage}[t]{0.48\textwidth}
\makeatletter\def\@captype{table}
\centering
\caption{\small{The results of HunyuanVideo on Videoalign and VisionReward trained with VideoAlign VQ\&MQ. "Baseline" denotes the original results of HunyuanVideo. We use the weighted sum of the probability for VisionReward-Video.}}
\label{table:video}
\scalebox{0.65}{
\begin{tabular}{ccccc}
\toprule
Benchmarks & VQ   & MQ  & TA & VisionReward   \\
\midrule
Baseline & 4.51 & 1.37 & \textbf{1.75} & 0.124 \\
Ours & \textbf{7.03} (+56\%) & \textbf{3.85} (+181\%) & 1.59 & \textbf{0.128}\\
\bottomrule
\end{tabular}}
\vspace{-5mm}
\end{minipage}

\subsection{Text-to-Image Generation}
\label{sec:t2i}
\begin{figure}[htbp]
  \centering
  \includegraphics[width=1.0\textwidth, height=0.20\textheight]{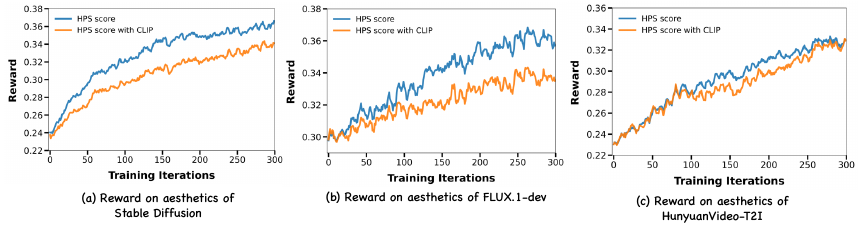}
  \caption{We visualize the reward curves of Stable Diffusion, FLUX.1-dev, and HunyuanVideo-T2I on HPS score from left to right. After applying CLIP score, the HPS score decreases, but the generated images become more natural (Figure \ref{fig:compar_1} in Appendix), and the CLIP score improves (Tables \ref{tab:sd} and \ref{tab:flux}).}
  \label{fig:curve_1}
  \vspace{-5mm}
\end{figure}

\textbf{Stable Diffusion.}
Stable Diffusion v1.4, a diffusion-based text-to-image generation framework, comprises three core components: a UNet architecture for iterative denoising, a CLIP-based text encoder for semantic conditioning, and a variational autoencoder (VAE) for latent space modeling. As demonstrated in Table~\ref{tab:sd} and Figure~\ref{fig:curve_1}(a), our proposed method, DanceGRPO, achieves a significant improvement in reward metrics, elevating the HPS score from 0.239 to 0.365, as well as the CLIP Score from 0.363 to 0.395.
We also take the metric like the Pick-a-Pic \cite{kirstain2023pick} and GenEval \cite{ghosh2023geneval} to evaluate our method. The results confirm the effectiveness of our method.
Moreover, as shown in Table~\ref{table:sd_compar}, our method exhibits the best performance in terms of metrics compared to other methods. We implement DPO as an online version following \cite{zhang2024onlinevpo}. 

Building on insights from rule-based reward models such as DeepSeek-R1, we conduct preliminary experiments with a binary reward formulation. By thresholding the continuous HPS reward at 0.28—assigning (CLIP score at 0.39) a value of 1 for rewards above this threshold and 0 otherwise, we construct a simplified reward model. Figure~\ref{fig:binary}(a) illustrates that DanceGRPO effectively adapts to this discretized reward distribution, despite the inherent simplicity of the thresholding approach. 
These results indicate the effectiveness of binary reward models in visual generation tasks.
In the future, we will also strive to explore more powerful rule-based visual reward models, for example, making judgments through a multimodal large language model.


\textit{Best-of-N Inference Scaling.} We explore sample efficiency through Best-of-N inference scaling using Stable Diffusion, as detailed in Section~\ref{sec:approach_dancegrpo}. By training the model on subsets of 16 samples (with the top 8 and bottom 8 rewards) selected from progressively larger pools (16, 64, and 256 samples per prompt), we evaluate the impact of sample curation on convergence dynamics about Stable Diffusion. Figure~\ref{fig:curve_3}(a) reveals that Best-of-N scaling substantially accelerates convergence. This underscores the utility of strategic sample selection in reducing training overhead while maintaining performance. For alternative approaches, such as tree search or greedy search, we defer their systematic integration to future research.

\textbf{FLUX.}
FLUX.1-dev is a flow-based text-to-image generation model that advances the state-of-the-art across multiple benchmarks, leveraging a more complex architecture than Stable Diffusion. To optimize performance, we integrate two reward models: HPS score and CLIP score. As illustrated in Figure~\ref{fig:curve_1}(b) and Table~\ref{tab:flux}, the proposed training paradigm achieves significant improvements across all reward metrics. 

\textbf{HunyuanVideo-T2I.}
HunyuanVideo-T2I is a text-to-image adaptation of the HunyuanVideo framework, reconfigured by reducing the number of latent frames to one. This modification transforms the original video generation architecture into a flow-based image synthesis model. We further optimize the system using the publicly available HPS-v2.1 model, a human-preference-driven metric for visual quality. As demonstrated in Figure~\ref{fig:curve_1}(c), this approach elevates the mean reward score from about 0.23 to 0.33, reflecting enhanced alignment with human aesthetic preferences.

\begin{figure}[htbp]
  \centering
  \includegraphics[width=1.0\textwidth, height=0.20\textheight]{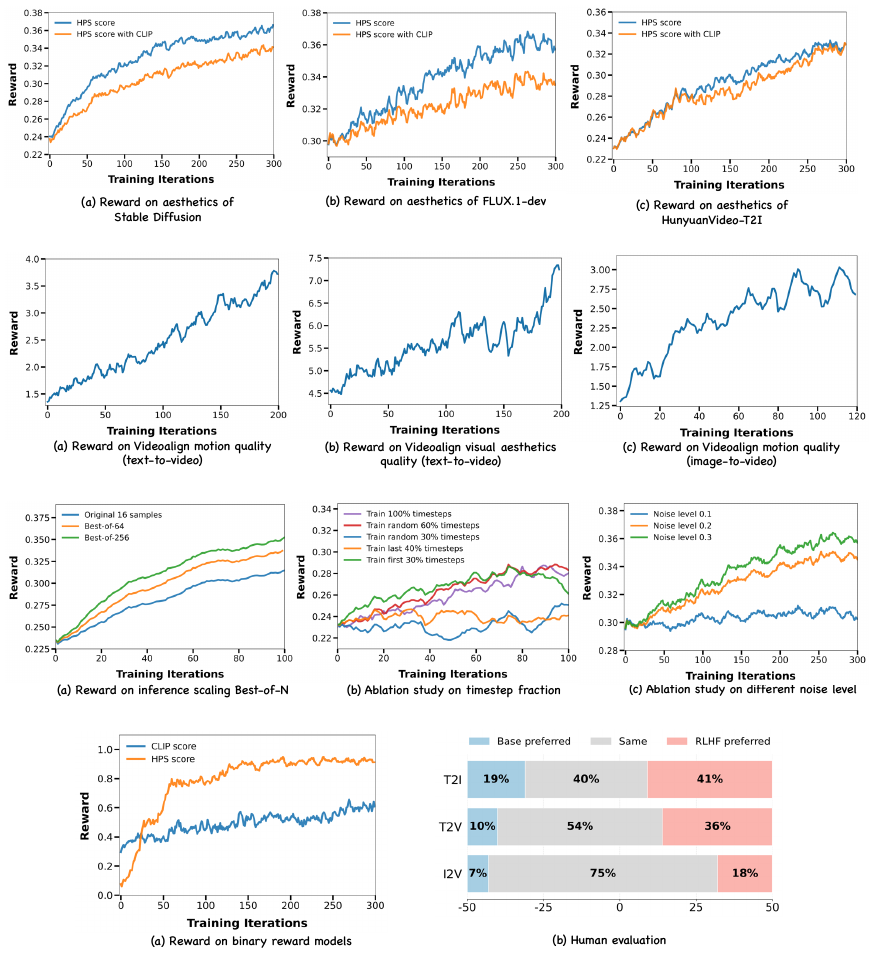}
  \caption{We visualize the training curves of motion quality and visual aesthetics quality on HunyuanVideo, motion quality on SkyReels-I2V.}
  \label{fig:curve_2}
  \vspace{-5mm}
\end{figure}

\subsection{Text-to-Video Generation}
\label{exp:t2v}
\textbf{HunyuanVideo.}
Optimizing text-to-video generation models presents significantly greater challenges compared to text-to-image frameworks, primarily due to elevated computational costs during training and inference, as well as slower convergence rates. In the pretraining protocol, we always adopt a progressive strategy: initial training focuses on text-to-image generation, followed by low-resolution video synthesis, and culminates in high-resolution video refinement. However, empirical observations reveal that relying solely on image-centric optimization leads to suboptimal video generation outcomes. To address this, our implementation employs training video samples synthesized at a resolution of 480×480 pixels, but we can visualize the samples with larger pixels.

Furthermore, constructing an effective video reward model for training alignment poses substantial difficulties. Our experiments evaluated several candidates: the Videoscore \cite{he2024videoscore} model exhibited unstable reward distributions, rendering it impractical for optimization, while Visionreward-Video \cite{xu2024visionreward}, a 29-dimensional metric, yielded semantically coherent rewards but suffered from inaccuracies across individual dimensions. Consequently, we adopted VideoAlign \cite{liu2025improving}, a multidimensional framework evaluating three critical aspects: visual aesthetics quality, motion quality, and text-video alignment. Notably, the text-video alignment dimension demonstrated significant instability, prompting its exclusion from our final analysis. We also increase the number of sampled frames per second for VideoAlign to improve the training stability. As illustrated in Figure~\ref{fig:curve_2}(a) and Figure~\ref{fig:curve_2}(b), our methodology achieves relative improvements of 56\% and \textbf{181\%} in visual and motion quality metrics, respectively. Extended qualitative results are provided in the Table~\ref{table:video}.

\subsection{Image-to-Video Generation}
\label{exp:i2v}
\textbf{SkyReels-I2V.} SkyReels-I2V represents a state-of-the-art open-source image-to-video (I2V) generation framework, established as of February 2025 at the inception of this study. Derived from the HunyuanVideo architecture, the model is fine-tuned by integrating image conditions into the input concatenation process. A central finding of our investigation is that I2V models exclusively allow optimization of motion quality, encompassing motion coherence and aesthetic dynamics, since visual fidelity and text-video alignment are inherently constrained by the attributes of the input image rather than the parametric space of the model. Consequently, our optimization protocol leverages the motion quality metric from the VideoAlign reward model, achieving a \textbf{118\%} relative improvement in this dimension as shown in Figure~\ref{fig:curve_2}(c). We must enable the CFG-training to ensure the sampling quality during the RLHF training process.

\begin{figure}[htbp]
  \centering
  \includegraphics[width=1.0\textwidth, height=0.22\textheight]{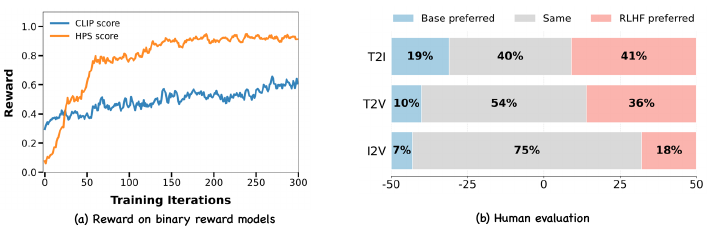}
  \caption{(a) We visualize the training curves of binary rewards. (b) We show the human evaluation results using FLUX (T2I), HunyuanVideo (T2V), and SkyReel (I2V), respectively.}
  \label{fig:binary}
  \vspace{-5mm}
\end{figure}

\subsection{Human Evaluation}
We present the results of our human evaluation, conducted using in-house prompts and reference images. For text-to-image generation, we evaluate FLUX on 240 prompts. For text-to-video generation, we assess HunyuanVideo on 200 prompts, and for image-to-video generation, we test SkyReels-I2V on 200 prompts paired with their corresponding reference images. As shown in Figure~\ref{fig:binary}(b), human artists consistently prefer outputs refined with RLHF. More visualization results can be found in Figure~\ref{fig:overall_vis} in Appendix, and Appendix~\ref{sec:appendix_vis}.


\subsection{Ablation Study}
\label{sec:abaltion}

\begin{figure}[htbp]
  \centering
  \includegraphics[width=1.0\textwidth, height=0.17\textheight]{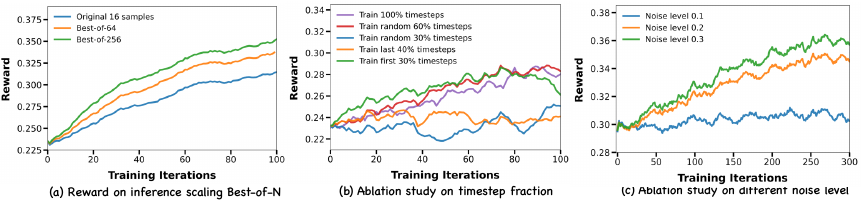}
  \caption{This figure shows the rewards on Best-of-N inference scaling, the ablation on timestep selection, and the ablation on noise level, respectively. While Best-of-N inference scaling consistently improves performance with more samples, it reduces sampling efficiency. Therefore, we leave Best-of-N as an optional extension.}
  \label{fig:curve_3}
  \vspace{-3mm}
\end{figure}

\textbf{Ablation on Timestep Selection.} As detailed in Section~\ref{sec:approach_dancegrpo}, we investigate the impact of timestep selection on the training dynamics of the HunyuanVideo-T2I model. We conduct an ablation study across three experimental conditions: (1) training exclusively on the first 30\% of timesteps from noise, (2) training on randomly sampled 30\% of timesteps, (3) training on the final 40\% of timesteps before outputs, (4) training on randomly sampled 60\% of timesteps, and (5) training on sampled 100\% of timesteps. As shown in Figure~\ref{fig:curve_3}(b), empirical results indicate that the initial 30\% of timesteps are critical for learning foundational generative patterns, as evidenced by their disproportionate contribution to model performance. However, restricting training solely to this interval leads to performance degradation compared to full-sequence training, likely due to insufficient exposure to late-stage refinement dynamics.

To reconcile this trade-off between computational efficiency and model fidelity, we always implement a 40\% stochastic timestep dropout strategy during training. This approach randomly masks 40\% of timesteps across all phases while preserving temporal continuity in the latent diffusion process.  The findings suggest that strategic timestep subsampling can optimize resource utilization in flow-based generative frameworks.

\textbf{Ablation on Noise Level $\varepsilon_t$.}
We systematically investigate the impact of noise level $\varepsilon_t$ during training on FLUX. Our analysis reveals that reducing $\varepsilon_t$ leads to a significant performance degradation, as quantitatively demonstrated in Figure~\ref{fig:curve_3}(c). Notably, experiments with alternative noise decay schedules (e.g., those used in DDPM) show no statistically significant differences in output quality compared to our baseline configuration. Futhermore, the noise level larger than 0.3 sometimes leads to noisy images after RLHF training. 

\section{Related Work}
\textbf{Aligning Large Language Models.}
Large Language Models (LLMs) \cite{achiam2023gpt, chang2024survey, brown2020language, grattafiori2024llama, yang2024qwen2} are typically aligned with Reinforcement Learning from Human Feedback (RLHF) \cite{ouyang2022training, yang2024qwen2, grattafiori2024llama, lee2023rlaif, gao2023scaling, bai2022training}. RLHF involves training a reward function based on comparison data of model outputs to capture human preferences, which is then utilized in reinforcement learning to align the policy model. While some approaches leverage policy gradient methods, others focus on Direct Policy Optimization (DPO) \cite{rafailov2023direct}. Policy gradient methods have proven effective but are computationally expensive and require extensive hyperparameter tuning. In contrast, DPO offers a more cost-efficient alternative but consistently underperforms compared to policy gradient methods.

Recently, DeepSeek-R1 \cite{guo2025deepseek} demonstrated that the application of large-scale reinforcement learning with formatting and result-only reward functions can guide LLMs toward the self-emergence of thought processes, enabling human-like complex chain-of-thought reasoning. This approach has achieved significant advantages in complex reasoning tasks, showcasing immense potential in advancing reasoning capabilities within Large Language Models. 

\textbf{Aligning Diffusion Models and Rectified Flows.} Diffusion models and rectified flows can also benefit significantly from alignment with human feedback, but the exploration remains primitive compared with LLMs. Key approaches in this area include: (1) Direct Policy Optimization (DPO)-style \cite{liang2024step, liu2025improving, xu2024visionreward, zhang2025diffusion, wallace2024diffusion} methods, (2) direct backpropagation with reward signals \cite{prabhudesai2024video}, such as ReFL \cite{xu2023imagereward}, and (3) policy gradient-based methods, including DPOK \cite{fan2023dpok} and DDPO \cite{black2023training}. However, production-level models predominantly rely on DPO and ReFL, as previous policy gradient methods have demonstrated instability when applied to large-scale settings. Our work addresses this limitation, providing a robust solution to enhance stability and scalability. We also hope our work offers insights into its potential to unify optimization paradigms across different modalities (e.g. image and text) \cite{ma2024janusflow, sun2024generative}.

\section{Conclusion and Future Work}
This work pioneers the integration of Group Relative Policy Optimization (GRPO) into visual generation, establishing DanceGRPO as a unified framework for enhancing diffusion models and rectified flows across text-to-image, text-to-video, and image-to-video tasks. By bridging the gap between language and visual modalities, our approach addresses critical limitations of prior methods, achieving superior performance through efficient alignment with human preferences and robust scaling to complex, multi-task settings. Experiments demonstrate substantial improvements in visual fidelity, motion quality, and text-image alignment. Future work will explore GRPO’s extension to multimodal generation, further unifying optimization paradigms across Generative AI.

\clearpage

\bibliographystyle{unsrt}
\bibliography{main}

\clearpage

\beginappendix

\section{Experimental Settings}
\label{sec:appendix_settings}
We provide detailed experimental settings in Table~\ref{tab:hyper-params}, which apply exclusively to training without classifier-free guidance (CFG). When enabling CFG, we configure one gradient update per iteration. Additionally, the sampling steps vary by model: we use 50 steps for Stable Diffusion, and 25 steps for FLUX and HunyuanVideo. 
\begin{table}[htbp]
\renewcommand{\arraystretch}{1.2}
\caption{\textbf{Our Hyper-paramters.}}
\centering
\begin{tabular}{cc}
\toprule
Learning rate & 1e-5 \\ 
Optimizer & AdamW  \\
Gradient clip norm  & 1.0   \\
Prompts per iteration & 32 \\
Images per prompt & 12 \\
Gradient updates per iteration & 4 \\
Clip range $\epsilon$ & 1e-4 \\
Noise level $\varepsilon_t$ & 0.3 \\
Timestep Selection $\tau$ & 0.6 \\
\bottomrule
\end{tabular}
\label{tab:hyper-params}
\end{table}

\section{More Analysis}
\label{sec:appendix_ana}

\subsection{Stochastic Interpolants}

The stochastic interpolant framework, introduced by \cite{albergo2303stochastic}, offers a unifying perspective on generative models like rectified flows and score-based diffusion models. It achieves this by constructing a continuous-time stochastic process that bridges any two arbitrary probability densities, $\rho_0$ and $\rho_1$.

In our work, we connect with a specific type of stochastic interpolant known as spatially linear interpolants, defined in Section 4 of \cite{albergo2303stochastic}. Given densities $\rho_0, \rho_1: \mathbb{R}^d \rightarrow \mathbb{R}_{\geq 0}$, a spatially linear stochastic interpolant process $x_t$ is defined as:
\begin{equation}
\mathbf{z}_t = \alpha(t) \mathbf{z}_0 + \beta(t) \mathbf{z}_1 + \gamma(t) \boldsymbol{\epsilon}, \quad t \in [0, 1],
\label{eq:si_linear}
\end{equation}
where $\mathbf{z}_0 \sim \rho_0$, $\mathbf{z}_1 \sim \rho_1$, and $\boldsymbol{\epsilon} \sim \mathcal{N}(0, I)$ is a standard Gaussian random variable independent of $\mathbf{z}_0$ and $\mathbf{z}_1$. The functions $\alpha, \beta, \gamma: [0, 1] \rightarrow \mathbb{R}$ are sufficiently smooth and satisfy the boundary conditions:
\begin{equation}
\alpha(0)=\beta(1)=1 , \quad \alpha(1)=\beta(0)=\gamma(0)=\gamma(1)=0,
\end{equation}
with the additional constraint that $\gamma(t) \geq 0$ for $t \in (0, 1)$. The term $\gamma(t)\mathbf{z}$ introduces latent noise, smoothing the path between densities.

Specific choices within this framework recover familiar models:
\begin{itemize}
\item \textbf{Rectified Flow (RF):} Setting $\gamma(t)=0$ (removing the latent noise), $\alpha(t) = 1-t$, and $\beta(t) = t$ yields the linear interpolation $\mathbf{z}_t = (1-t)\mathbf{z}_0 + t\mathbf{z}_1$ used in Rectified Flow \cite{liu2022flow, albergo2303stochastic}. The dynamics are typically governed by an ODE $\mathrm{d}\mathbf{z}_t = \mathbf{u}_t \mathrm{d}t$, where $\mathbf{u}_t$ is the learned velocity field.
\item \textbf{Score-Based Diffusion Models (SBDM):} The framework connects to SBDMs via one-sided linear interpolants (Section 4.4 of \cite{albergo2303stochastic}), where $\rho_1$ is typically Gaussian. The interpolant takes the form $\mathbf{z}_t = \alpha(t)\mathbf{z}_0 + \beta(t)\mathbf{z}_1$. The VP-SDE formulation \cite{song2020score} corresponds to choosing $\alpha(t) = \sqrt{1 - t^2}$ and $\beta(t) = t$ after a time reparameterization.
\end{itemize}

A pivotal insight is that: "The law of the interpolant $z_t$ at any time $t \in [0, 1]$ can be realized by many different processes, including an ODE and forward and backward SDEs whose drifts can be learned from data."

The stochastic interpolant framework provides a probability flow ODE for RF:
\begin{equation}
\mathrm{d}\mathbf{z}_t = \mathbf{u}_t \mathrm{d}t.
\label{ode:si_prob_flow}
\end{equation}
The backward SDE associated with the interpolant's density evolution is given by:
\begin{equation}
\mathrm{d}\mathbf{z}_t = \mathbf{b}_B(t, \mathbf{x}_t) \mathrm{d}t + \sqrt{2\epsilon(t)} \mathrm{d}\mathbf{z},
\label{sde:si_backward}
\end{equation}
where $\mathbf{b}_B(t, \mathbf{x}) = \mathbf{u}_t - \epsilon(t) \mathbf{s}(t, \mathbf{x})$ is the backward drift, $\mathbf{s}(t, \mathbf{x})$ is the score function, and $\epsilon(t) \ge 0$ is a tunable diffusion coefficient (noise schedule). If we set $\varepsilon_t = \sqrt{2\epsilon(t)}$, the backward SDE becomes:
\begin{equation}
\mathrm{d}\mathbf{z}_t = \left( \mathbf{u}_t - \frac{\varepsilon_t^2}{2} \nabla\log p_t(\mathbf{z}_t) \right) \mathrm{d}t + \varepsilon_t \mathrm{d}\mathbf{z},
\label{sde:si_backward_alt_eps}
\end{equation}
which is the same as \cite{gao2025diffusionmeetsflow}.

\subsection{Connections between Rectified Flows and Diffusion Models} 
We aim to demonstrate the equivalence between certain formulations of diffusion models and flow matching (specifically, stochastic interpolants) by deriving the hyperparameters of one model from the other.

The forward process of a diffusion model is described by an SDE:
\begin{equation} \label{eq:diffusion_forward}
\mathrm{d} \mathbf{z}_t = f_t \mathbf{z}_t \mathrm{d} t + g_t \mathrm{d} \mathbf{w} ,
\end{equation}
where $\mathrm{d} \mathbf{w}$ is a Brownian motion, and $f_t, g_t$ define the noise schedule.

The corresponding generative (reverse) process SDE is given by:
\begin{equation} \label{eq:diffusion_reverse}
\mathrm{d} \mathbf{z}_t = \left( f_t \mathbf{z}_t - \frac{1+ \eta_t^2}{2}g_t^2 \nabla \log p_t(\mathbf{z}_t) \right) \mathrm{d} t + \eta_t g_t \mathrm{d} \mathbf{w}  ,
\end{equation}
where $p_t(\mathbf{z}_t)$ is the marginal probability density of $\mathbf{z}_t$ at time $t$.

For flow matching, we consider an interpolant path between data $\mathbf{x} = \mathbf{z}_0$ and noise $\mathbf{\epsilon}$ (typically $\mathbf{\epsilon} \sim \mathcal{N}(0, \mathbf{I})$):
\begin{equation} \label{eq:interpolant_path}
\mathbf{z}_t = \alpha_t \mathbf{x} + \sigma_t \mathbf{\epsilon}.
\end{equation}
This path satisfies the ODE:
\begin{equation} \label{eq:flow_ode}
\mathrm{d}\mathbf{z}_t = \mathbf{u}_t \mathrm{d}t, \quad \text{where } \mathbf{u}_t = \dot{\alpha}_t \mathbf{x} + \dot{\sigma}_t \mathbf{\epsilon}.
\end{equation}
This can be generalized to a stochastic interpolant SDE:
\begin{equation} \label{eq:stochastic_interpolant_sde}
\mathrm{d} \mathbf{z}_t = (\mathbf{u}_t - \frac{1}{2} \varepsilon_t^2 \nabla \log p_t(\mathbf{z}_t)) \mathrm{d} t + \varepsilon_t \mathrm{d} \mathbf{w} .
\end{equation}

The core idea is to match the marginal distributions $p_t(\mathbf{z}_t)$ generated by the forward diffusion process Eq.\eqref{eq:diffusion_forward} with those implied by the interpolant path Eq.\eqref{eq:interpolant_path}. We will derive $f_t$ and $g_t$ from this requirement, and then relate the noise terms of the generative SDEs Eq.\eqref{eq:diffusion_reverse} and Eq.\eqref{eq:stochastic_interpolant_sde} to find $\eta_t$.

\textbf{Deriving $f_t$ by Matching Means.} From Eq.\eqref{eq:interpolant_path}, assuming $\mathbf{x} = \mathbf{z}_0$ is fixed and $E[\mathbf{\epsilon}] = \mathbf{0}$, the mean of $\mathbf{z}_t$ is $E[\mathbf{z}_t] = \alpha_t \mathbf{x}$. The mean $\mathbf{m}_t = E[\mathbf{z}_t]$ of the process Eq.\eqref{eq:diffusion_forward} starting from $\mathbf{z}_0 = \mathbf{x}$ satisfies the ODE $\frac{\mathrm{d} \mathbf{m}_t}{\mathrm{d} t} = f_t \mathbf{m}_t$. We require $\mathbf{m}_t = \alpha_t \mathbf{x}$ for all $t$. Substituting into the mean ODE:
\begin{equation}
    \frac{\mathrm{d}}{\mathrm{d} t} (\alpha_t \mathbf{x}) = f_t (\alpha_t \mathbf{x}), \qquad \dot{\alpha}_t \mathbf{x} = f_t \alpha_t \mathbf{x}.
\end{equation}
Assuming this holds for any $\mathbf{x}$ and $\alpha_t \neq 0$, we divide by $\alpha_t \mathbf{x}$:
\begin{equation}
    f_t = \frac{\dot{\alpha}_t}{\alpha_t}
\end{equation}
Using the identity $\frac{\mathrm{d}}{\mathrm{d} t} \log(y) = \dot{y} / y$, we get:
\begin{equation} \label{eq:f_t_result}
    f_t = \partial_t \log(\alpha_t)
\end{equation}

\textbf{Deriving $g_t^2$ by Matching Variances.} From Eq.\eqref{eq:interpolant_path}, assuming $\mathbf{x}$ is fixed and $Var(\mathbf{\epsilon}) = \mathbf{I}$ (identity matrix for standard Gaussian noise), the variance (covariance matrix) of $\mathbf{z}_t$ is $Var(\mathbf{z}_t) = Var(\alpha_t \mathbf{x} + \sigma_t \mathbf{\epsilon}) = \sigma_t^2 Var(\mathbf{\epsilon}) = \sigma_t^2 \mathbf{I}$. Let $V_t = \sigma_t^2$ be the scalar variance magnitude. The variance $V_t = \text{Tr}(Var(\mathbf{z}_t))/d$ for the process Eq.\eqref{eq:diffusion_forward} evolves according to the Lyapunov equation:
 $   \frac{\mathrm{d} V_t}{\mathrm{d} t} = 2 f_t V_t + g_t^2$
(Here, $g_t^2$ represents the magnitude of the noise injection rate). We require $V_t = \sigma_t^2$. Substitute $V_t = \sigma_t^2$ and $f_t = \dot{\alpha}_t / \alpha_t$ into the variance evolution equation:
\begin{equation}
    \frac{\mathrm{d}}{\mathrm{d} t} (\sigma_t^2) = 2 \left( \frac{\dot{\alpha}_t}{\alpha_t} \right) \sigma_t^2 + g_t^2, \qquad
    2 \sigma_t \dot{\sigma}_t = 2 \frac{\dot{\alpha}_t}{\alpha_t} \sigma_t^2 + g_t^2.
\end{equation}
Solving for $g_t^2$:
\begin{equation}
        g_t^2 = 2 \sigma_t \dot{\sigma}_t - 2 \frac{\dot{\alpha}_t}{\alpha_t} \sigma_t^2 
              = \frac{2}{\alpha_t} ( \alpha_t \sigma_t \dot{\sigma}_t - \dot{\alpha}_t \sigma_t^2 ) 
              = \frac{2 \sigma_t}{\alpha_t} ( \alpha_t \dot{\sigma}_t - \dot{\alpha}_t \sigma_t )
\end{equation}
    Using the quotient rule for differentiation, $\partial_t (\sigma_t / \alpha_t) = \frac{\alpha_t \dot{\sigma}_t - \dot{\alpha}_t \sigma_t}{\alpha_t^2}$, which implies $\alpha_t \dot{\sigma}_t - \dot{\alpha}_t \sigma_t = \alpha_t^2 \partial_t (\sigma_t / \alpha_t)$. Then we get:
\begin{equation}
        g_t^2 = \frac{2 \sigma_t}{\alpha_t} \left( \alpha_t^2 \partial_t \left( \frac{\sigma_t}{\alpha_t} \right) \right) 
              = 2 \alpha_t \sigma_t \partial_t \left( \frac{\sigma_t}{\alpha_t} \right)
\end{equation}
    Thus, we have:
    \begin{equation} \label{eq:g_t_squared_result}
        g_t^2 = 2 \alpha_t \sigma_t \partial_t \left( \frac{\sigma_t}{\alpha_t} \right)
    \end{equation}

\textbf{Deriving $\eta_t$ by Matching Noise Terms in Generative SDEs.}
We compare the coefficients of the Brownian motion term ($\mathrm{d} \mathbf{w}$) in the reverse diffusion SDE Eq.\eqref{eq:diffusion_reverse} and the stochastic interpolant SDE Eq.\eqref{eq:stochastic_interpolant_sde}.
The diffusion coefficient (magnitude of the noise term) is $D_{\text{diff}} = \eta_t g_t$. The diffusion coefficient is $D_{\text{int}} = \varepsilon_t$. To match the noise structure in these specific SDE forms, we set $D_{\text{diff}} = D_{\text{int}}$: $\eta_t g_t = \varepsilon_t$.
    Solving for $\eta_t$ (assuming $g_t \neq 0$):$\eta_t = \frac{\varepsilon_t}{g_t}$.
Substitute $g_t = \sqrt{g_t^2}$ using the result from Eq.\eqref{eq:g_t_squared_result}:
\begin{equation} \label{eq:eta_t_result}
    \eta_t = \frac{\varepsilon_t}{\sqrt{2 \alpha_t \sigma_t \partial_t (\sigma_t / \alpha_t)}}
\end{equation}

\textbf{Summary of Results.} By requiring the forward diffusion process Eq.\eqref{eq:diffusion_forward} to match the marginal mean and variance of the interpolant path Eq.\eqref{eq:interpolant_path} at all times $t$, we derived:
\begin{equation}
    f_t = \partial_t \log(\alpha_t) , \quad g_t^2 = 2 \alpha_t \sigma_t \partial_t (\sigma_t / \alpha_t) , \quad \eta_t = \varepsilon_t / (2 \alpha_t \sigma_t \partial_t (\sigma_t / \alpha_t))^{1/2} .
\end{equation}
These relationships establish the equivalence between the parameters of the two frameworks under the specified conditions.

\section{Classifier-Free Guidance (CFG) Training}
\label{sec:appendix_cfg}
Classifier-Free Guidance (CFG) \cite{ho2022classifier} is a widely adopted technique for generating high-quality samples in conditional generative modeling. However, in our settings, integrating CFG into training pipelines introduces instability during optimization. To mitigate this, we empirically recommend disabling CFG during the sampling phase for models with high sample fidelity, such as HunyuanVideo and FLUX, as it reduces gradient oscillation while preserving output quality.

For CFG-dependent models like SkyReels-I2V and Stable Diffusion, where CFG is critical for reasonable sample quality, we identify a key instability: training exclusively on the conditional objective leads to divergent optimization trajectories. This necessitates the joint optimization of both conditional and unconditional outputs, effectively doubling VRAM consumption due to dual-network computations. Morever, we propose reducing the frequency of parameter updates per training iteration. For instance, empirical validation shows that limiting updates to one per iteration significantly enhances training stability for SkyReels-I2V, with minimal impact on convergence rates.

\section{Advantages over DDPO and DPOK} 
\label{sec:appendix_overddpo}
Our approach differs from prior RL-based methods for text-to-image diffusion models (e.g., DDPO, DPOK) in three key aspects: (1) We employ a GRPO-style objective function, (2) we compute advantages within prompt-level groups rather than globally, (3) we ensure noise consistency across samples from the same prompt, (4) we generalize these improvements beyond diffusion models by applying them to rectified flows and scaling to video generation tasks.

\section{Inserting DDPO into Rectified Flow SDEs}
\label{sec:ddpo}
\begin{figure}[htbp]
  \centering
  \includegraphics[width=1.0\textwidth, height=0.22\textheight]{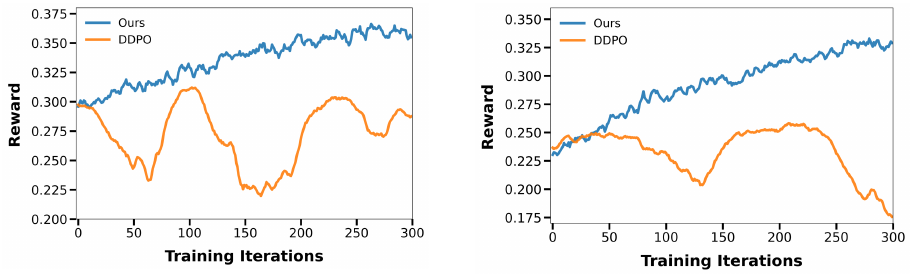}
\caption{We visualize the results of DDPO and Ours. DDPO always diverges when applied to rectified flow SDEs.}
  \label{fig_app:ddpo}
\end{figure}

We also insert DDPO-style objective function into rectified flow SDEs, but it always diverges, as shown in Figure~\ref{fig_app:ddpo}, which demonstrates the superiority of DanceGRPO.

\section{More Visualization Results}
\label{sec:appendix_vis}
\begin{figure}[htbp]
  \centering
  \includegraphics[width=1.0\textwidth, height=0.57\textheight]{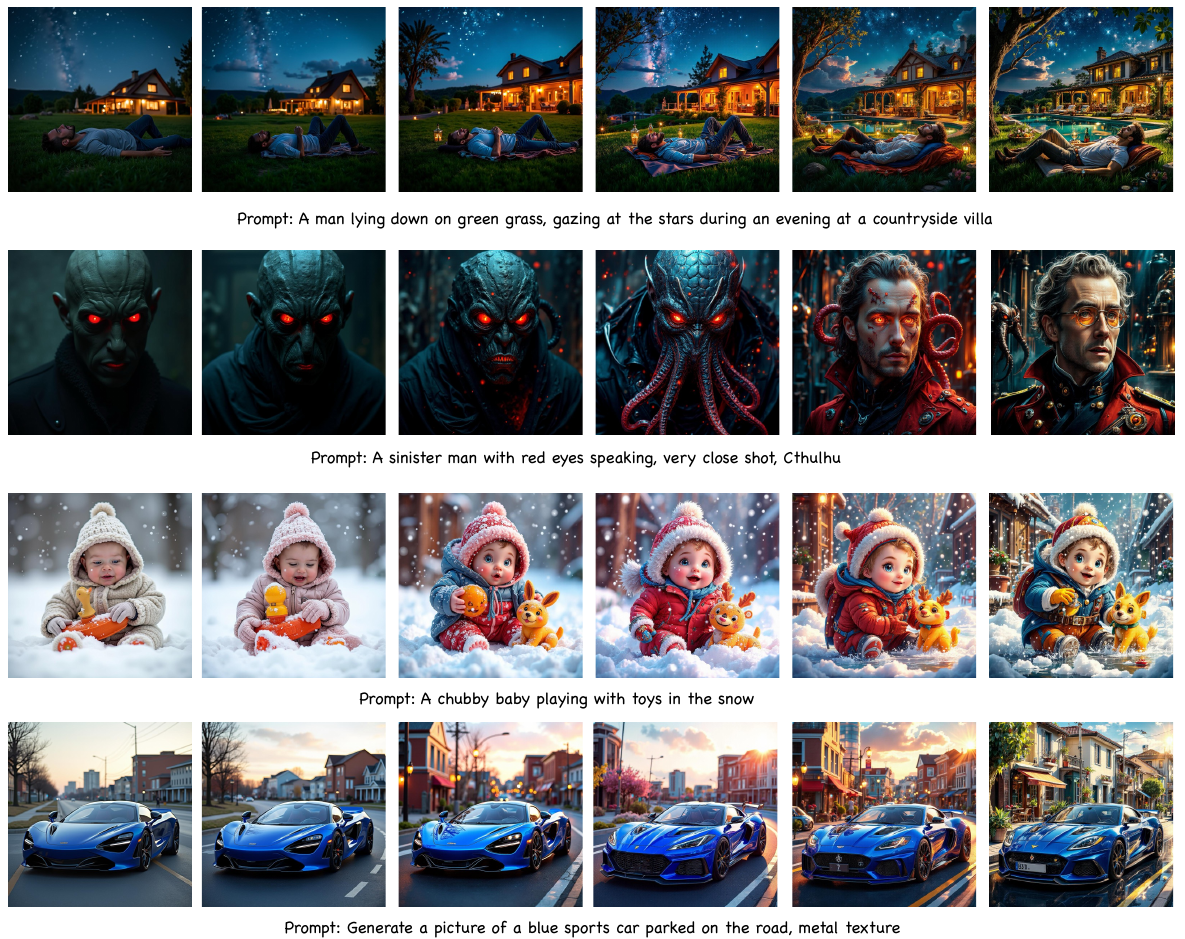}
\caption{We visualize the results by selecting FLUX optimized with the HPS score at iterations 0, 60, 120, 180, 240, and 300. The optimized outputs tend to exhibit brighter tones and richer details. However, incorporating CLIP score regularization is crucial, as demonstrated in Figures~\ref{fig_app:flux_1},~\ref{fig_app:flux_2}, and~\ref{fig_app:flux_3}.}
  \label{fig_app:flux_process}
\end{figure}

\begin{figure}[!htb]
  \centering
  \includegraphics[width=1.0\textwidth, height=0.23\textheight]{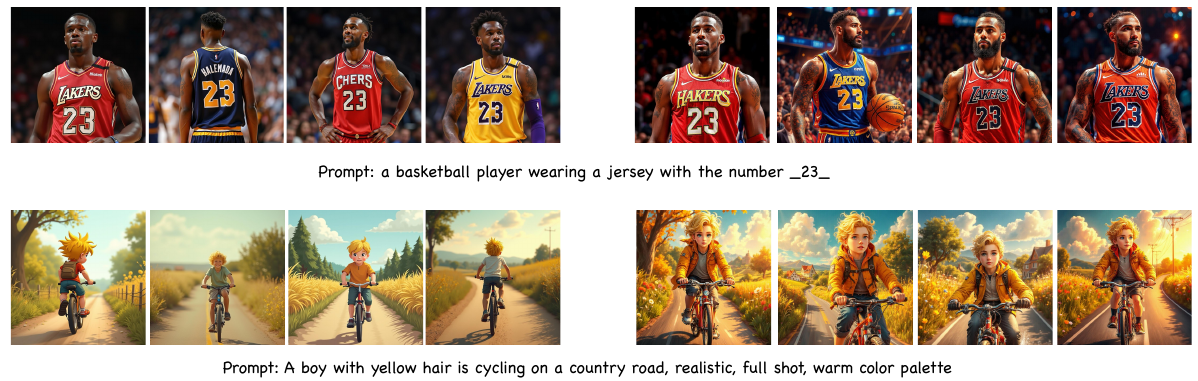}
\caption{Visualization of the diversity of the model before and after RLHF. Different seed tends to generate similar images after RLHF.}
  \label{fig_app:flux_diversity}
\end{figure}

\begin{figure}[!htb]  \centering
    \includegraphics[width=1.0\textwidth, height=0.67\textheight]{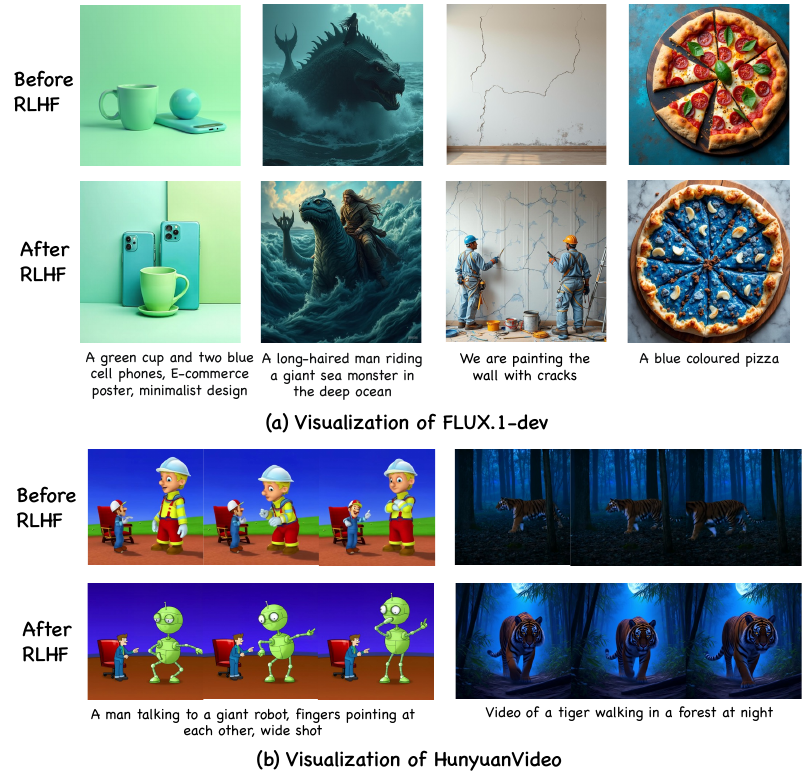}
  \caption{\textbf{Overall visualization.} We visualize the results before and after RLHF of FLUX and HunyuanVideo.}
  \label{fig:overall_vis}
\end{figure}

\begin{figure}[htbp]
  \centering
  \includegraphics[width=0.9\textwidth, height=0.93\textheight]{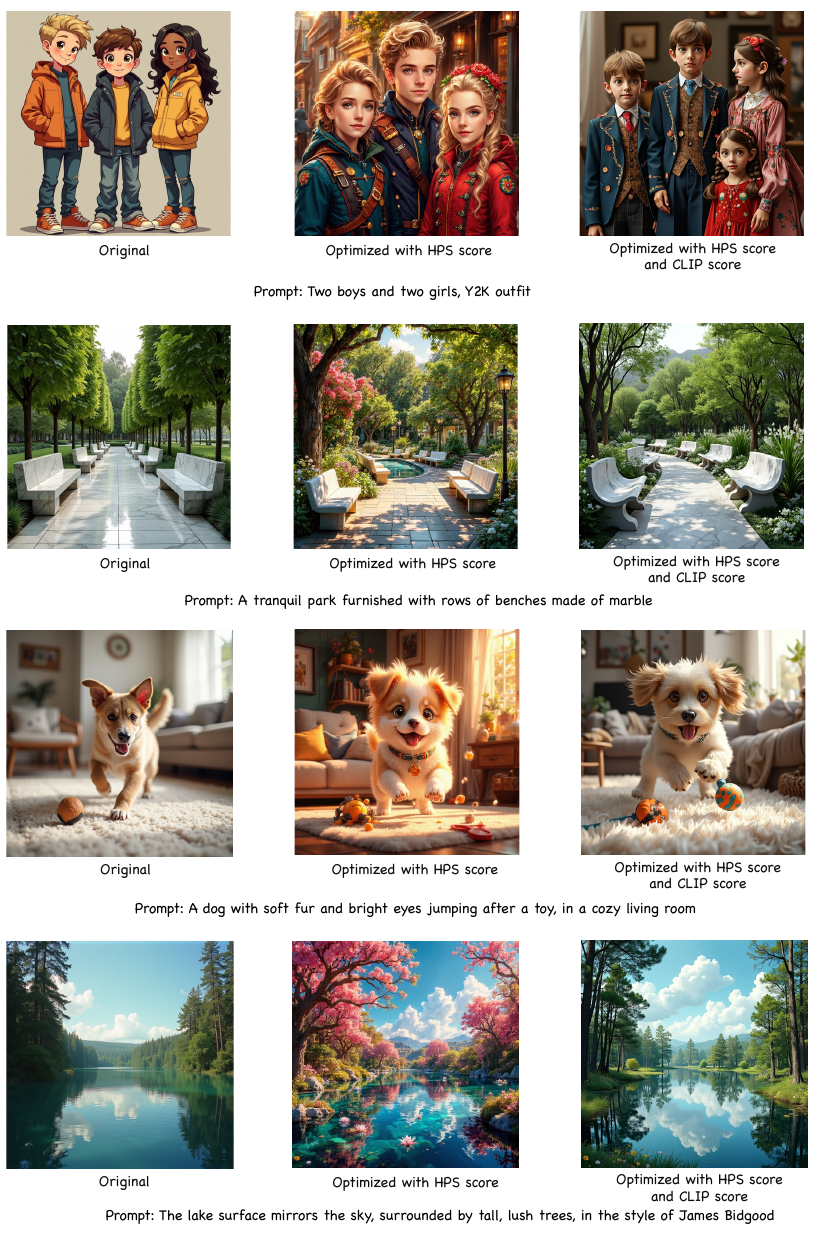}
  \caption{We present the original outputs of FLUX, alongside optimizations driven solely by the HPS score and those enhanced by both the HPS and CLIP scores.}
  \label{fig_app:flux_1}
\end{figure}

\begin{figure}[htbp]
  \centering
  \includegraphics[width=0.9\textwidth, height=0.93\textheight]{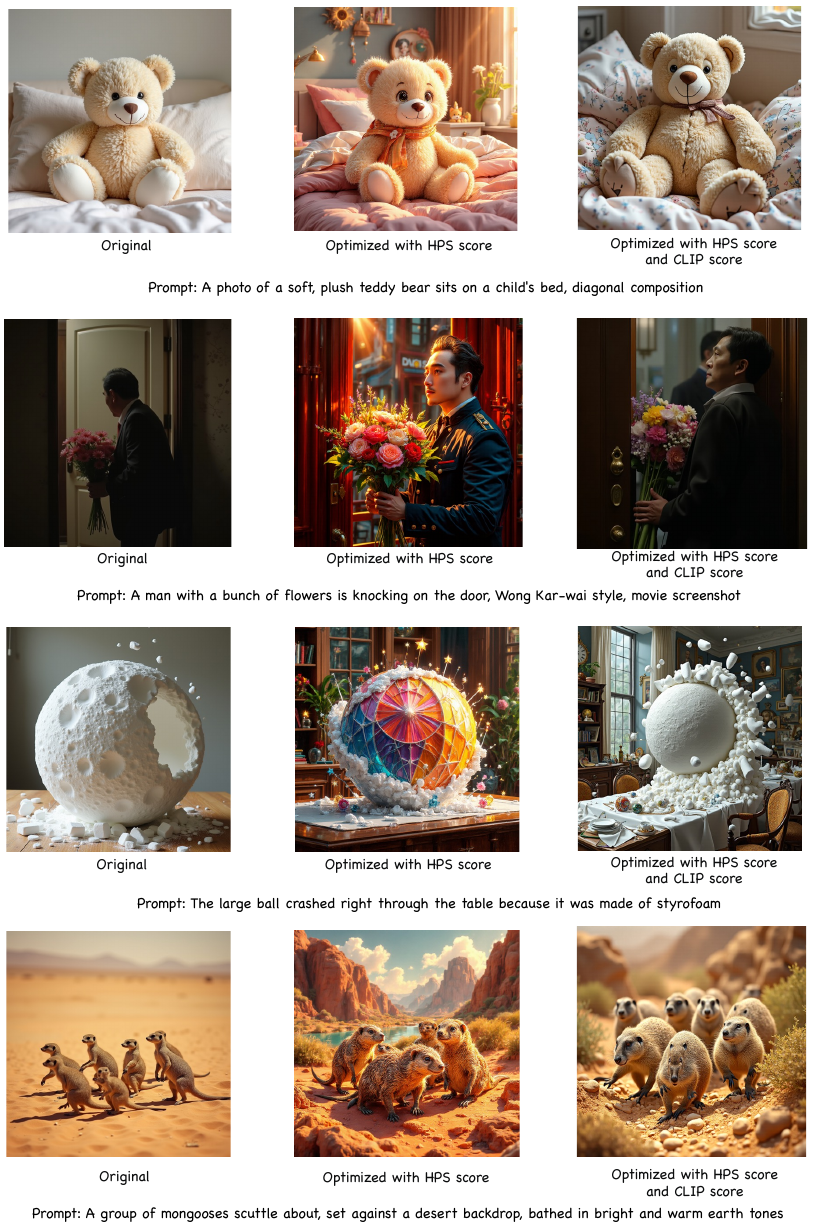}
  \caption{We present the original outputs of FLUX, alongside optimizations driven solely by the HPS score and those enhanced by both the HPS and CLIP scores.}
  \label{fig_app:flux_2}
\end{figure}

\begin{figure}[htbp]
  \centering
  \includegraphics[width=0.9\textwidth, height=0.93\textheight]{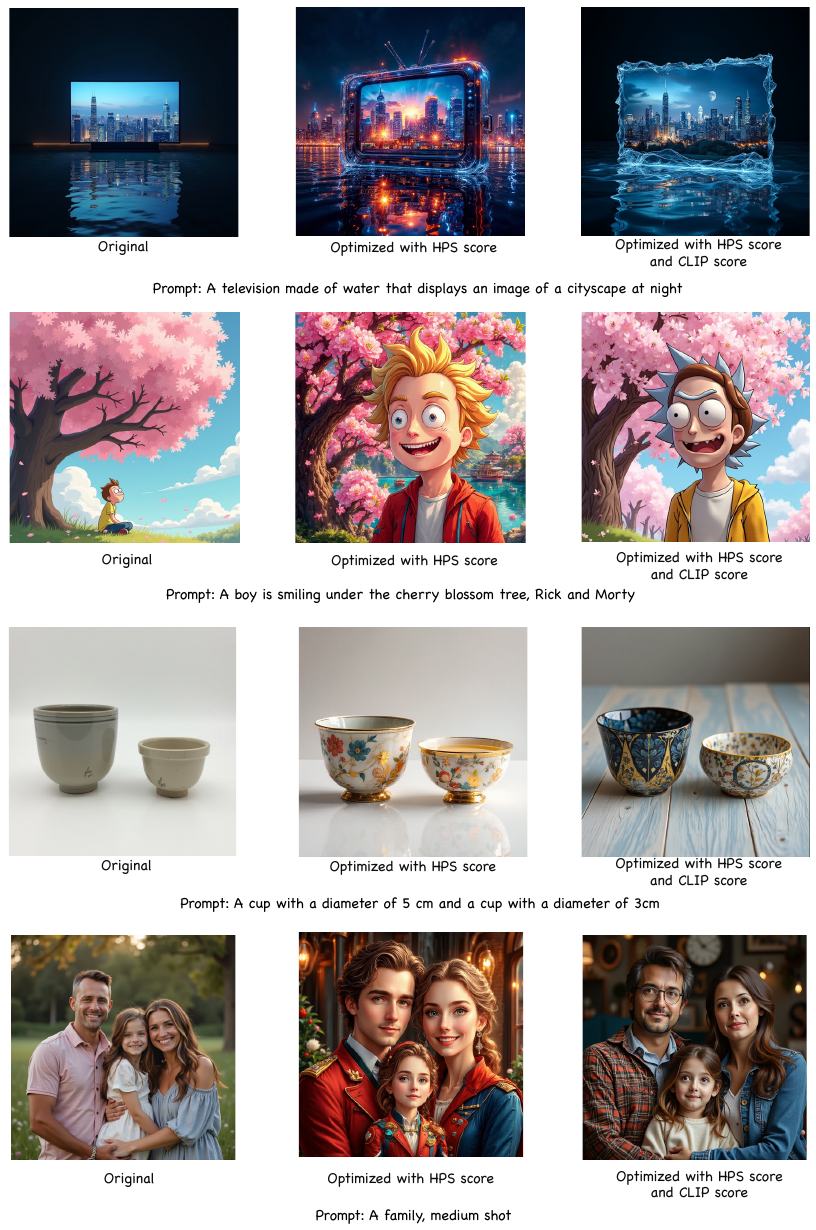}
  \caption{We present the original outputs of FLUX, alongside optimizations driven solely by the HPS score and those enhanced by both the HPS and CLIP scores.}
  \label{fig_app:flux_3}
\end{figure}

\begin{figure}[htbp]
  \centering
  \includegraphics[width=0.9\textwidth, height=0.9\textheight]{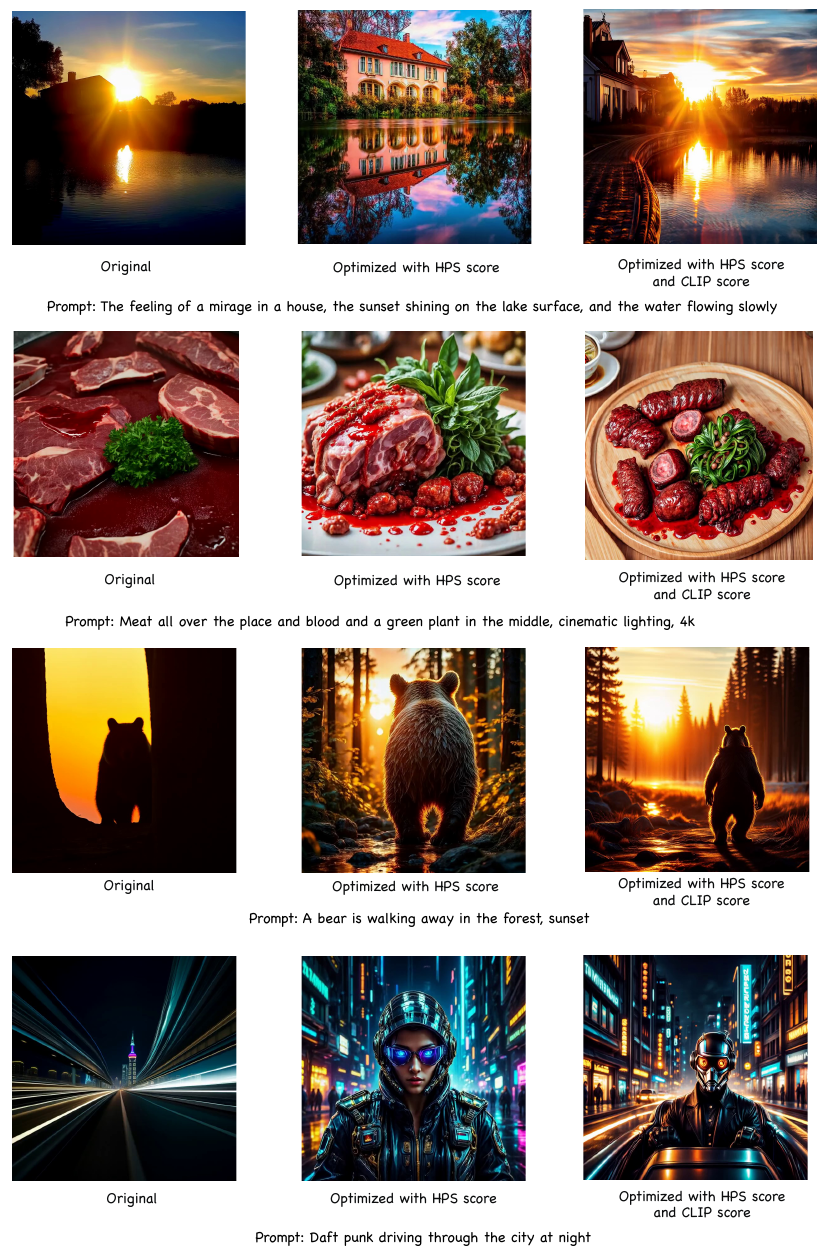}
  \caption{We present the original outputs of HunyuanVideo-T2I, alongside optimizations driven solely by the HPS score and those enhanced by both the HPS and CLIP scores.}
  \label{fig_app:hunyuant2i}
\end{figure}

\begin{figure}[htbp]
  \centering
  \includegraphics[width=0.9\textwidth, height=0.9\textheight]{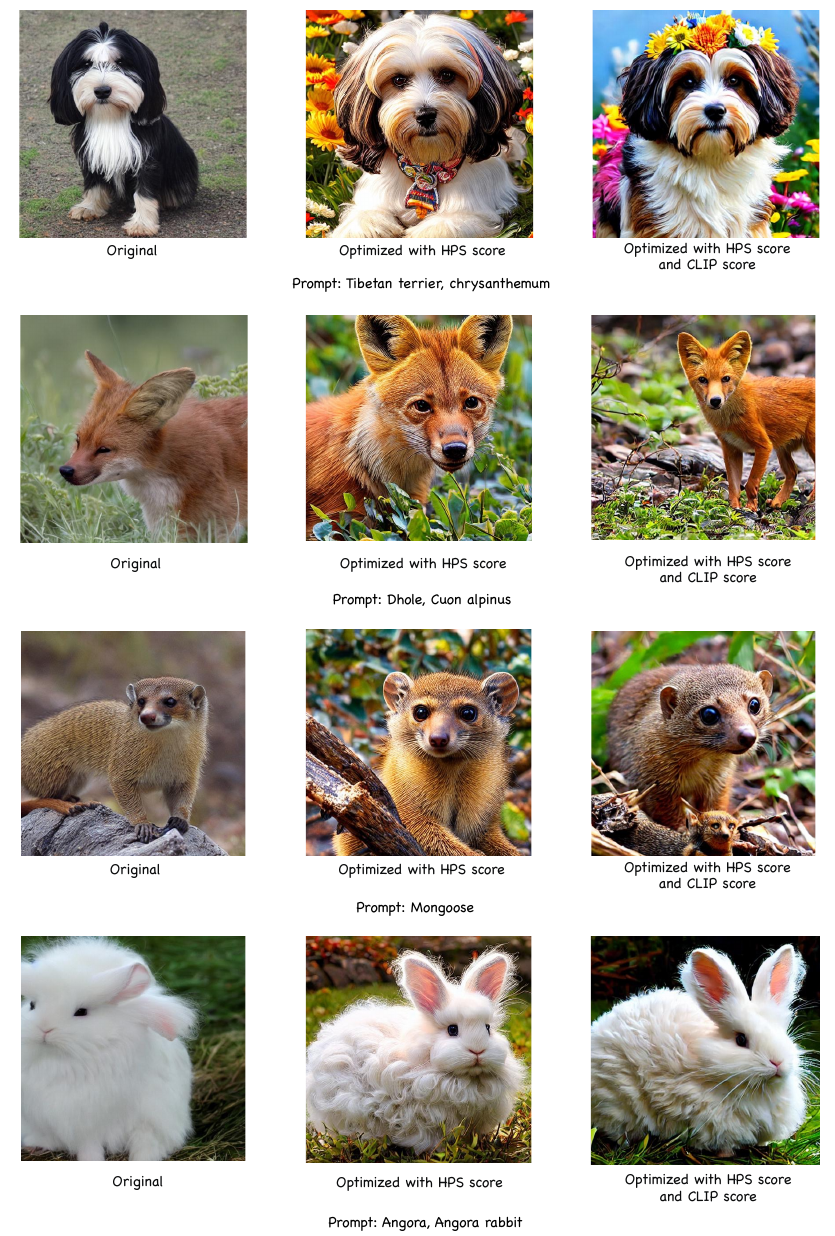}
  \caption{We present the original outputs of Stable Diffusion, alongside optimizations driven solely by the HPS score and those enhanced by both the HPS and CLIP scores.}
  \label{fig_app:sd1}
\end{figure}

\begin{figure}[htbp]
  \centering
  \includegraphics[width=1.0\textwidth, height=0.77\textheight]{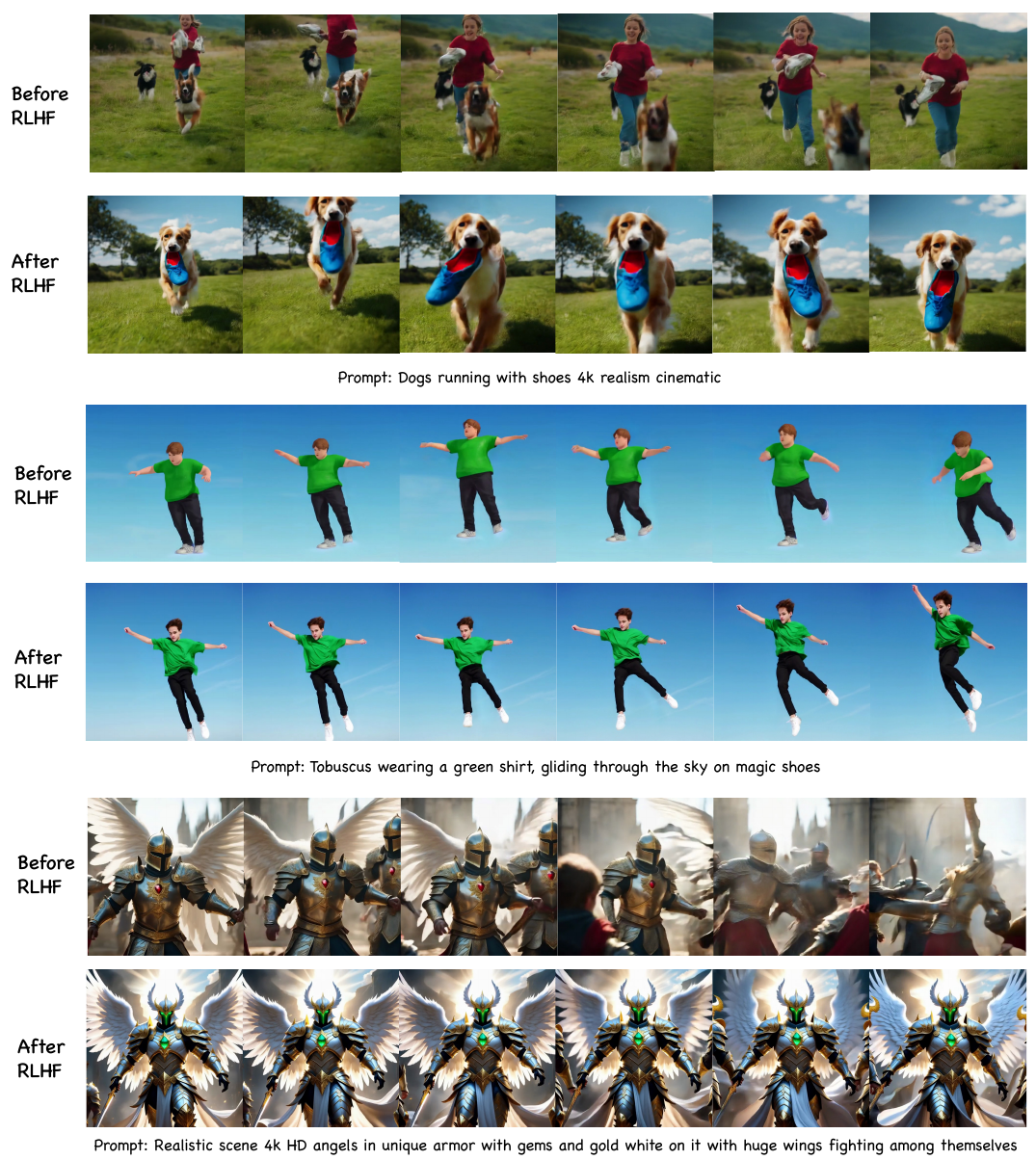}
  \caption{Visualization results of HunyuanVideo.}
  \label{fig_app:t2v1}
\end{figure}

\begin{figure}[htbp]
  \centering
  \includegraphics[width=1.0\textwidth, height=0.77\textheight]{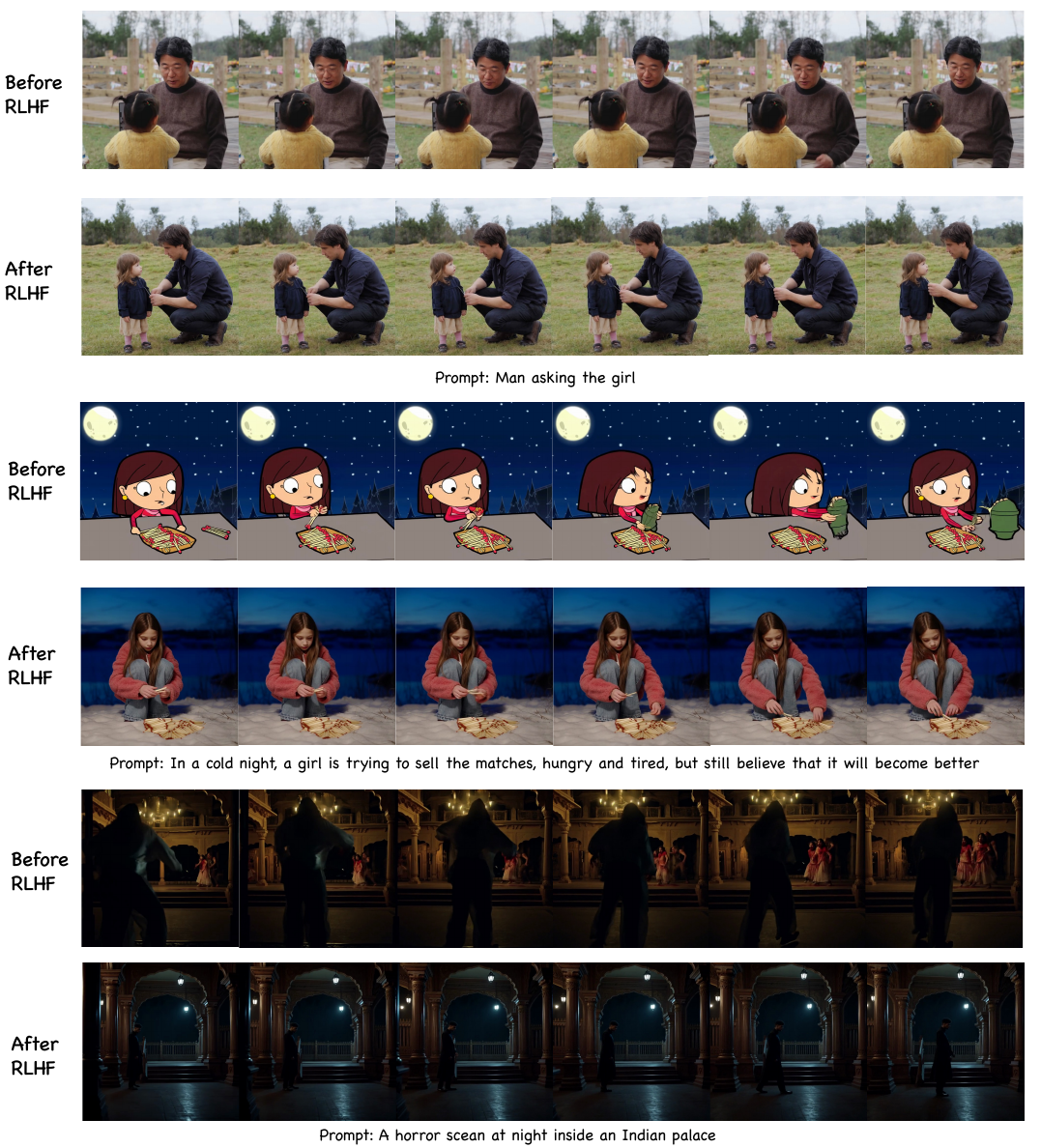}
  \caption{Visualization results of HunyuanVideo.}
  \label{fig_app:t2v2}
\end{figure}

\begin{figure}[htbp]
  \centering
  \includegraphics[width=1.0\textwidth, height=0.77\textheight]{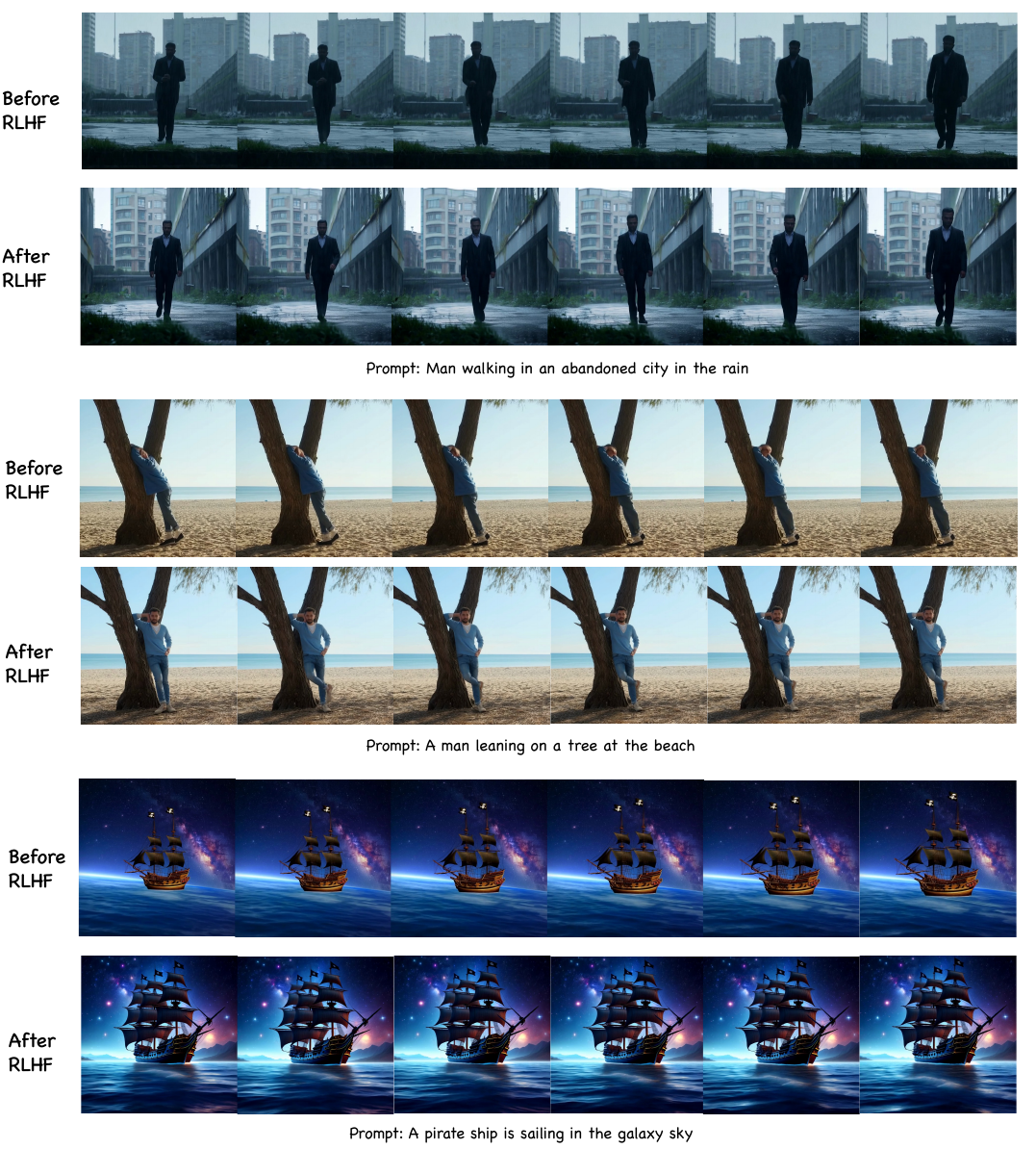}
  \caption{Visualization results of HunyuanVideo.}
  \label{fig_app:t2v3}
\end{figure}

We provide more visualization results on FLUX, Stable Diffusion, and HunyuanVideo as shown in Figure~\ref{fig_app:flux_process},~\ref{fig_app:flux_diversity},~\ref{fig:overall_vis},~\ref{fig_app:flux_1},~\ref{fig_app:flux_2},~\ref{fig_app:flux_3}, ~\ref{fig_app:hunyuant2i}, ~\ref{fig_app:sd1},~\ref{fig_app:t2v1},~\ref{fig_app:t2v2}, and ~\ref{fig_app:t2v3}.

\begin{figure}[!htb]
  \centering
  \includegraphics[width=1.0\textwidth, height=0.13\textheight]{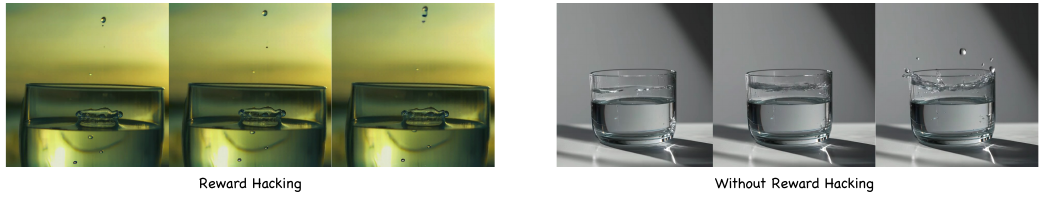}
\caption{Visualization of the results of reward hacking (with different initialization noise) and without reward hacking (with the same initialization noise) on HunyuanVideo. Prompt: A splash of water in a clear glass, with sparkling, clear radiant reflections, sunlight, sparkle}
  \label{fig_app:reward_hacking}
\end{figure}

\begin{figure}[htbp]
  \centering
  \includegraphics[width=0.9\textwidth, height=0.22\textheight]{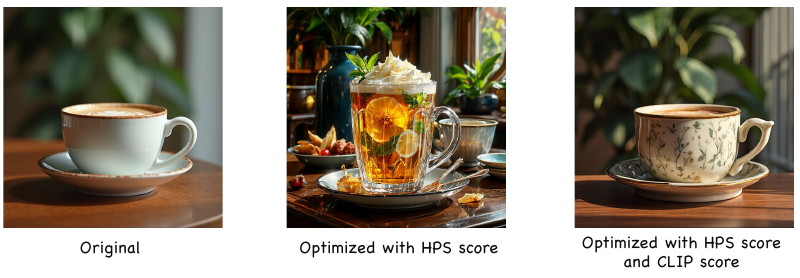}
  \caption{This figure demonstrates the impact of the CLIP score. The prompt is "A photo of cup". We find that the model trained solely with HPS-v2.1 rewards tends to produce unnatural ("oily") outputs, while incorporating CLIP scores helps maintain more natural image characteristics.}
  \label{fig:compar_1}
\end{figure}

\begin{figure}[htbp]
  \centering
  \includegraphics[width=1.0\textwidth, height=0.84\textheight]{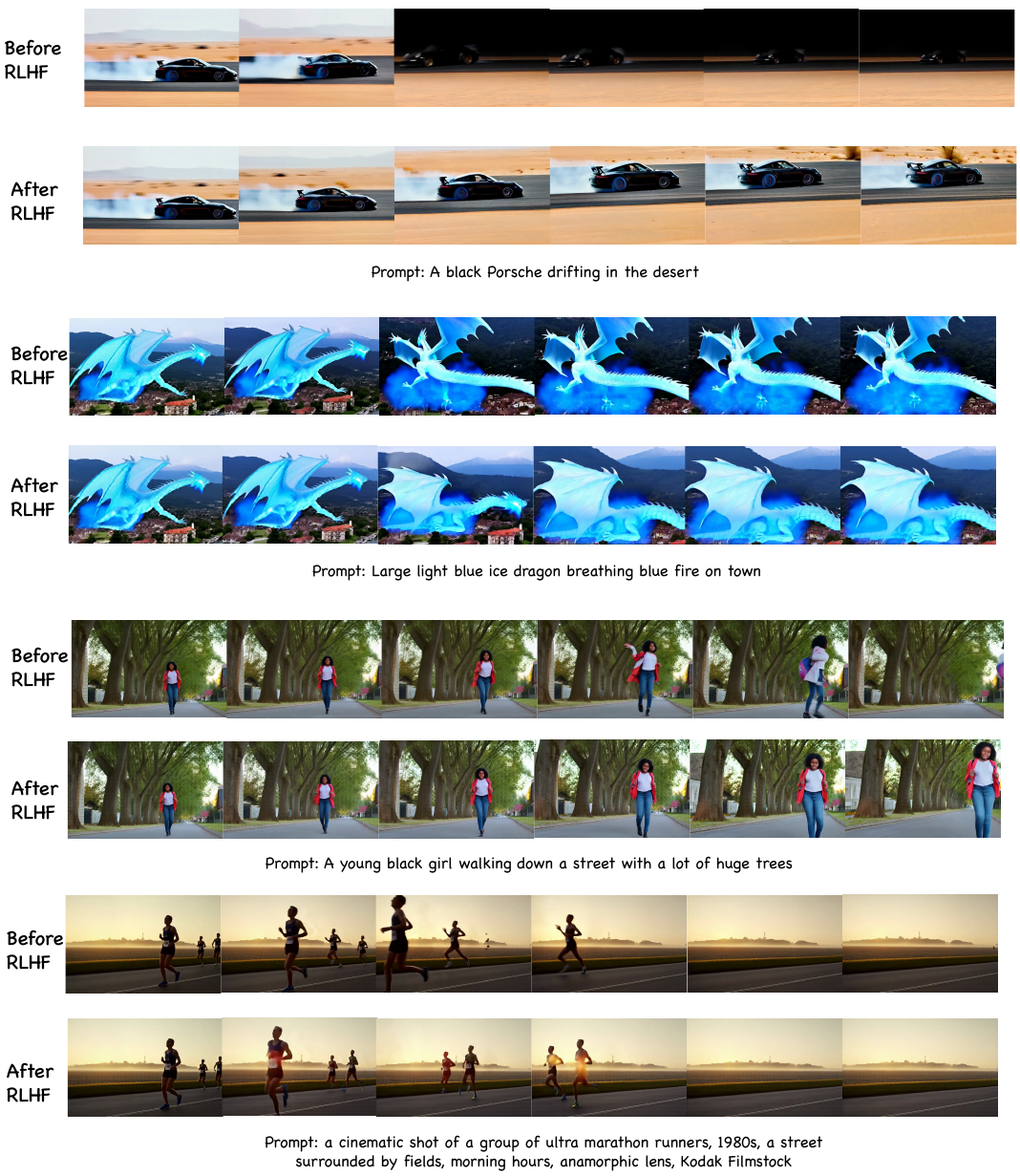}
  \caption{Visualization results of SkyReels-I2V.}
  \label{fig_app:i2v}
\end{figure}

\end{document}